\documentclass[journal]{IEEEtai}

\usepackage[colorlinks,urlcolor=blue,linkcolor=blue,citecolor=blue]{hyperref}

\usepackage{color,array}

\usepackage{xcolor,soul,framed} 

\colorlet{shadecolor}{yellow}
\usepackage[pdftex]{graphicx}
\graphicspath{{./Images/pdf/}{./Images/jpeg/}{./Images/}}
\DeclareGraphicsExtensions{.pdf,.jpeg,.png}
\usepackage[cmex10]{amsmath}
\usepackage{array}
\usepackage{mdwmath}
\usepackage{mdwtab}
\usepackage{eqparbox}
\usepackage{url}
\usepackage{comment}
\usepackage{amsmath,amssymb} 
\usepackage{color}
\usepackage{epsfig}
\usepackage{epstopdf}
\usepackage{subcaption}
\usepackage{comment}
\usepackage{bm}
\usepackage{float}
\usepackage{algorithm} 
\usepackage{algorithmic}  
\usepackage[algo2e]{algorithm2e}
\usepackage{soul}
\usepackage[switch]{lineno}
\usepackage[backend=bibtex,bibstyle=ieee,citestyle=numeric-comp]{biblatex}
\bibliography{refs}

\colorlet{BLACK}{black}

\begin{document}

\title{Histogram Layers for Texture Analysis}

\author{Joshua Peeples,~\IEEEmembership{Student Member,~IEEE,}
      Weihuang Xu,~\IEEEmembership{Student Member,~IEEE,}
      Alina Zare,~\IEEEmembership{Senior Member,~IEEE}

\thanks{Manuscript received June 28, 2021; revised October 29, 2021 and November 27, 2021; accepted December 10, 2021.}
\thanks{J.Peeples and W. Xu is a PhD candidate in the Department of Electrical and Computer Engineering, University of Florida, Gainesville, FL, 32608 USA (e-mail: jpeeples@ufl.edu, weihuang.xu@ufl.edu).}
  \thanks{W. Xu is a PhD candidate in the Department of Electrical and Computer Engineering, University of Florida, Gainesville, FL, 32608 USA (e-mail: weihuang.xu@ufl.edu).}%
  \thanks{A. Zare is a Professor in the Department of Electrical and Computer Engineering, University of Florida, Gainesville, FL, 32608 USA (e-mail: azare@ece.ufl.edu).}
  \thanks{This material is based upon work supported by the National Science Foundation Graduate Research Fellowship under Grant No. DGE-1842473. The views and opinions of the authors expressed herein do not necessarily state or reflect those of the United States Government or any agency thereof. The authors acknowledge University of Florida Research Computing for providing computational resources and support that have contributed to the research results reported in this publication: \url{http://researchcomputing.ufl.edu}}}

\markboth{Accetped to Journal of IEEE Transactions on Artificial Intelligence}
{Peeples \MakeLowercase{\textit{et al.}}: Histogram Layers for Texture Analysis}

\maketitle

\begin{abstract}
An essential aspect of texture analysis is the extraction of features that describe the distribution of values in local, spatial regions. We present a \textit{localized} histogram layer for artificial neural networks. Instead of computing global histograms as done previously, the proposed histogram layer directly computes the local, spatial distribution of features for texture analysis, and parameters for the layer are estimated during backpropagation. We compare our method to state-of-the-art texture encoding methods such as: the Deep Encoding Pooling Network, Deep Texture Encoding Network, Fisher Vector convolutional neural network, and Multi-level Texture Encoding and Representation. We used three material/texture datasets: (1) the Describable Texture Dataset; (2) an extension of the ground terrain in outdoor scenes dataset; and (3) a subset of the Materials in Context dataset. Results indicate that the inclusion of the proposed histogram layer improves performance. The source code for the histogram layer is publicly available \footnote{\url{https://github.com/GatorSense/Histogram_Layer}}.
\end{abstract}

\begin{IEEEImpStatement}
As currently constructed, convolutional neural networks (CNN) cannot directly model texture information in data; and texture serves as a powerful descriptor that can be used in variety of tasks. We propose new histogram layers to characterize the distribution of features in a CNN. Our proposed method fuses aspects of handcrafted texture features and automated feature learning of deep learning. Histogram layers are expected to not only capture texture information to improve performance, but also serve as explainable and interpretable components for deep learning models.
\end{IEEEImpStatement}


\section{Introduction}

\begin{figure}[t]
	\begin{center}
		\includegraphics[draft=false,width=1\linewidth]{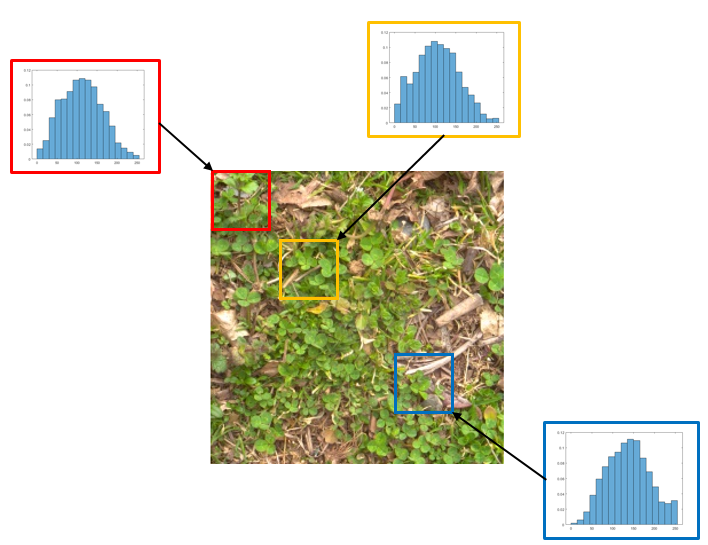}
	\end{center}
	\caption{This is an example image of grass from GTOS-mobile \cite{xue2018deep}. The image contains other textures and not only grass. Local histograms can distinguish portions of the image containing pure grass (top two histograms) or a mixture of other textures (bottom histogram). Integrating a histogram layer in deep neural networks will assist in estimating the data distribution to improve texture analysis. The histograms shown here are the distribution of intensity values from the red, green, and blue channels. Each histogram contains the aggregated intensity values (over the three color channels) in the corresponding image portion.}
	\label{fig:RBF_Histograms}
\end{figure}

\IEEEPARstart{T}{exture} analysis is a crucial component in many applications including autonomous vehicles \cite{curio1999walking}, automated medical diagnosis \cite{castellano2004texture}, and explosive hazard detection \cite{anderson2012combination}. The concept of texture is easily discernible for humans, but there is no agreed definition within the computer vision community \cite{tuceryan1993texture,liu2019bow}. Generally, variations in the definition of texture arise because of differences in the application being studied (\textit{i.e.}, texture characteristics that are more informative vary across application areas) \cite{tuceryan1993texture,liu2019bow}. Yet, most agree that one common component of texture analysis relies on characterizing the spatial distribution of intensity and/or feature values as shown in Figure {\ref{fig:RBF_Histograms}}.  
	
A number of handcrafted features have been developed with successful application to texture-dependent computer vision problems; however, the process to design these features can be difficult. Feature engineering is an expensive process in terms of labor, computation, and time. It often requires significant domain knowledge and expertise, as well. Additionally, these features often rely on empirically determining the best parameters for each descriptor, resulting in an increase of computation and time.  For example, histogram-based features, such as histogram of oriented gradients (HOG) \cite{dalal2005histograms} and local binary patterns (LBP) \cite{ojala2002multiresolution}, have been extensively studied and used in texture-based applications \cite{lowe1999object,frigui2008detection,ojala1994performance,yang2010bag}. In both HOG and LBP feature sets, spatial distributions of feature values are used to characterize and distinguish textures.  Furthermore, the distributions are summarized using histograms. 

\begin{figure*}[t]
	\begin{subfigure}{.55\textwidth}{
			\includegraphics[draft=false,width=\textwidth]{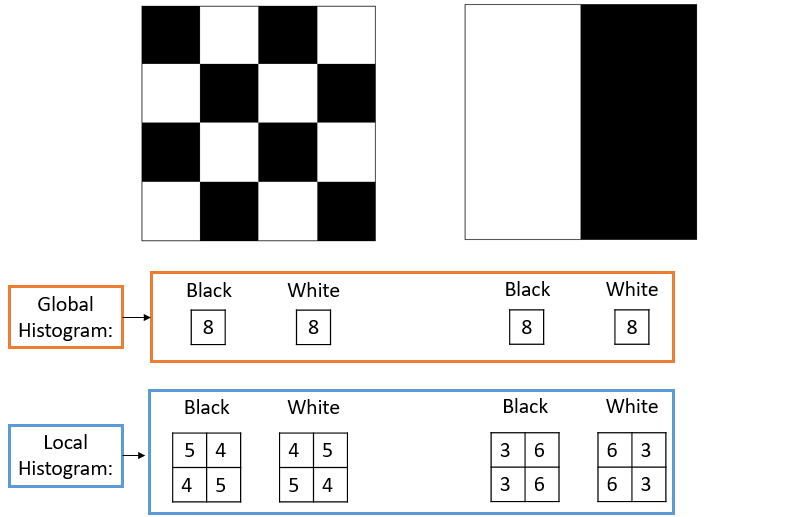}
			\caption{Local vs global histogram}
			\label{fig:Toy_Hist}
		}
	\end{subfigure} 
	\hspace{1mm}
	\begin{subfigure}{.425\textwidth}{
			\includegraphics[draft=false,width=\textwidth]{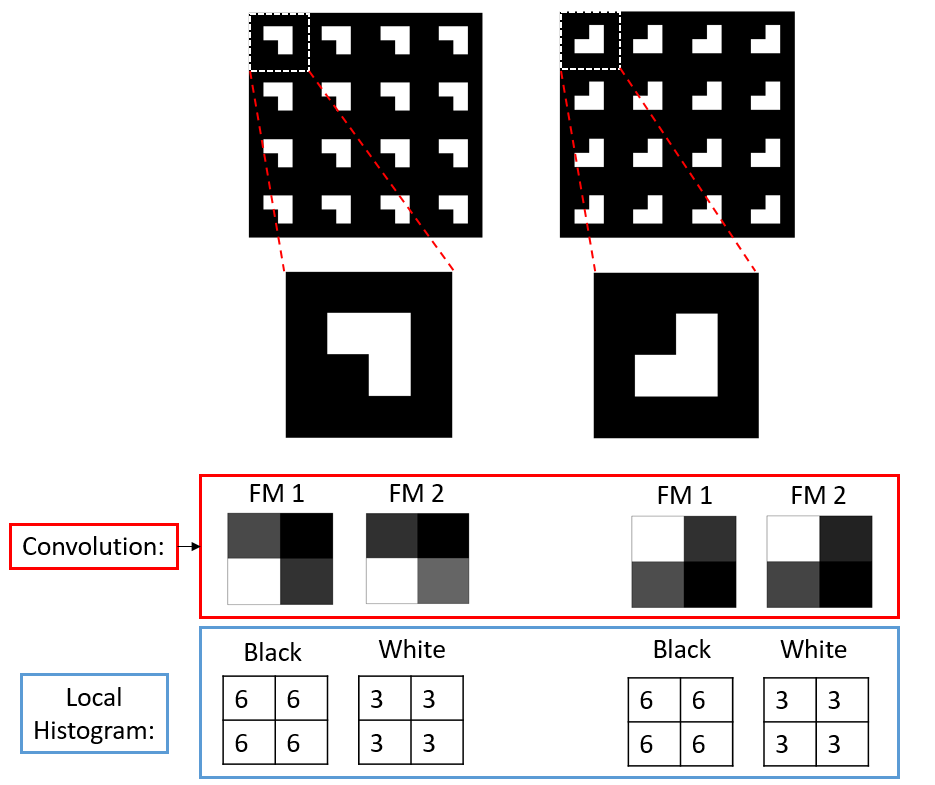}
			\caption{Local histogram vs convolution}
			\label{fig:Toy_conv}
		}
	\end{subfigure}
	\caption{Toy 4$\times$4 images showing the disadvantages of a global histogram (\ref{fig:Toy_Hist}) and a convolution operation (\ref{fig:Toy_conv}). The local histograms and convolutions use a kernel size of $3 \times 3$ to produce $2 \times 2$ feature maps. On the left in \ref{fig:Toy_Hist}, the two images are distinct textures. If a global histogram is computed, the distribution of white and black pixels is equivalent, resulting in no discrimination between the two texture types. On the right in \ref{fig:Toy_conv}, convolution operations are sensitive to image transformations such as rotations. The two textures shown are the same, but applying filters from a convolutional neural network such as ResNet18 in local areas results in different feature maps (FM). However, a local histogram provides some rotational invariance and learns the same local distribution of pixels for each image.}
	\label{fig:Toy_example}
\end{figure*} 

In recent work, handcrafted feature extraction has often been substituted with automated feature learning using deep learning to address some of the issues associated with designing features. Deep learning has often outperformed approaches that couple hand-designed feature extraction with classification, segmentation and object recognition  \cite{he2016deep,krizhevsky2012imagenet,liang2015recurrent,long2015fully}. Despite the success of deep learning, some works have shown empirically and theoretically that traditional features perform better, or are comparable to, that of deep learning methods in texture analysis \cite{basu2016theoretical,basu2018deep,cavalin2017review,liu2017local}. Additionally, these deep learning models cannot model the distribution of values in regions, a factor that is essential for texture analysis \cite{tuceryan1993texture}. Deep architectures require more layers resulting in more parameters to characterize the spatial distribution of features in a convolutional neural network (CNN), as opposed to using a histogram directly.    

The proposed solution, which is a \textit{histogram layer} for artificial neural networks (ANNs), automates the feature learning process while simultaneously modelling the distribution of features. The histogram layer is a tool to integrate and utilize the strengths of both handcrafted features and deep learning, in order to maximize texture analysis performance. Histograms are an effective and efficient approach to aggregate information. The selection of the bin centers and widths are crucial for the feature representation. Instead of manually determining these parameters, these parameters are estimated through backpropagation. Radial basis functions (RBFs) are used as the histogram-binning operation to allow for the gradient information to flow through the network \cite{sedighi2017histogram}.  The unique aspects of this work are:

\begin{itemize}
	\item The first localized histogram layer for texture analysis which maintains spatial context
	\item Similar to Wang et al.'s histogram layer \cite{wang2016learnable}, the bin centers and widths of the histogram are estimated through backpropagation
	\item As shown with RBFs \cite{sedighi2017histogram}, our approach is robust to ambiguity and outliers through the use of ``soft'' bin assignments.
\end{itemize}

\section{Related Work}
	\subsection{Deep Learning for Texture Analysis} \label{sect:DL_texture}
	Deep learning has been successfully used to achieve state-of-the-art performance for texture applications \cite{liu2019bow,cavalin2017review}. Attempts to combine neural and traditional features into deep learning architectures have shown success \cite{paul2016combining,wu2016multi,wang2014mitosis,rivera2018densenet}, but the traditional features can not be updated through this process. Also, some have tried to emulate handcrafted features via the network design \cite{bruna2013invariant,chan2015pcanet,malof2018improving} but have run into issues including high computational costs and a decrease in texture analysis performance \cite{liu2019bow}.  Another approach for texture analysis is to aggregate the features extracted by pre-trained models through ``encoding layers" \cite{cimpoi2015deep,zhang2017deep,song2017locally,xue2018deep}. As noted by Liu et al. \cite{liu2019texture}, these ``encoding layers" have primarily focused on transfer learning approaches for CNNs, but convolutional features are sensitive to image transformations \cite{liu2019texture,liu2016evaluation} such as rotations as shown in Figure \ref{fig:Toy_example}. There are current efforts to improve the robustness of CNNs through rotation invariant (or equivariant) convolutions \mbox{\cite{marcos2016learning,andrearczyk2018rotational,laptev2016ti}}. However, these approaches possibly may still have issues dealing with intra-class variations of rotated images. Additionally, these approaches may limit the generalization ability of the model due to constraints on the architecture \mbox{\cite{kang2021rotation}}. The proposed histogram layer will be more robust than previous methods` ``encoding layers" due to ``soft'' binning assignments that are less sensitive to ambiguity and outliers in the feature maps. The proposed histogram layer also can be jointly trained with the convolutional layers to influence the features learned by the network; thus, there is no need for strong constraints as in the case of some rotation invariant CNN models. 

	\subsection{Pooling Operations}
	Common components of deep learning frameworks are pooling layers such as max pooling (\textit{i.e.},  captures the highest feature values) and average pooling (\textit{i.e.}, computes the mean of each descriptor). Generally, pooling layers are used to aggregate feature information over patches of the image. Pooling layers provide several advantages such as generalization, reduced computational costs, and improved performance \cite{Goodfellow-et-al-2016,akhtar2019interpretation}. However, these pooling layers make assumptions about the data that are not optimal for every dataset \cite{akhtar2019interpretation}. For example, some data types (such as synthetic aperture sonar imagery) are plagued with difficult-to-remove speckle noise \cite{abu2018robust}. The use of min or max pooling will tend to propagate noise values as opposed to more informative values. 
	
	Also, several pooling operations (\textit{e.g.}, max pooling) only backpropagate the error through certain locations resulting in a saturation issue that slows learning \cite{yu2014mixed}. The proposed histogram layer will retain the advantages of standard pooling operations but will learn the bin centers and widths necessary to aggregate the features of the data throughout the data distribution.  The proposed histogram layer will also be robust to outliers in the data. If a value is far from each bin center, the contribution of the outlier will be negligible. Additionally, the proposed histogram layer also provides normalization of the features because the contribution of each descriptor for each bin is in the range of $[0,1]$.
	
	\subsection{Influence of Architecture Design}
	The selection of certain architecture elements will introduce inductive biases for ANNs \mbox{\cite{neyshabur2015search,xu2021positional,phuong2020inductive,bronstein2021geometric}}. Multi-layer perceptrons (MLPs) can be used to theoretically approximate any function {\cite{pinkus1999approximation}}, but this is not practical due to the curse of dimensionality \mbox{\cite{bronstein2021geometric}}. Additionally, MLPs do not have a mechanism to account for local relationships in the data \mbox{\cite{bronstein2021geometric,Goodfellow-et-al-2016}}. Convolutional layers were introduced to account for this limitation of MLPs, as the convolution operation is translation equivariant \mbox{\cite{bronstein2021geometric,Goodfellow-et-al-2016}} and uses shared weights to extract features from neighboring inputs.   Another inductive bias that can be introduced is permutation invariance \mbox{\cite{bronstein2021geometric}}. Examples of permutation invariant models are graph neural networks \mbox{\cite{bronstein2021geometric,xu2018powerful}}, Deep Sets \mbox{\cite{zaheer2017deep}}, and Transformers \mbox{\cite{vaswani2017attention}}. A graph is used to describe relationships between collection of elements \mbox{\cite{xu2018powerful,bronstein2021geometric}}. Images are a special case of graphs that have fixed ordering and structure \mbox{\cite{wu2020comprehensive,bronstein2021geometric}}. A graph, $\mathcal{G}$, is defined as a collection of nodes $\mathcal{V}$ and edges $\mathcal{E} \subseteq \mathcal{V} \times \mathcal{V}$ between node pairs: $\mathcal{G} = (\mathcal{V},\mathcal{E})$. Below is a brief introduction of \mbox{\cite{bronstein2021geometric}}, the concept of permutation invariance through \textit{sets}, a special type of graphs without edges. Given a set of nodes and $d$ features associated with each of the $n$ nodes, one can stack the rows to generate a $n \times d$ matrix $\mathbf{X} = (\mathbf{x}_1,...,\mathbf{x}_n)^T$. A permutation matrix, $P \in \mathbb{R}^{n \times n}$, denotes the reordering of the rows of $\mathbf{X}$. A function, $f$, is permutation invariant if the constraint of $f(\mathbf{PX}) = f(\mathbf{X})$ is satisfied. For finite sets, given suitable mappings $\psi$ and $\phi$, every permutation-invariant function can be decomposed into Equation {\ref{eqn:perm}}, where $\psi$ is a function applied independently to each node's feature vector and $\phi$ is function applied to the aggregated sum of the outputs from $f$ \mbox{\cite{zaheer2017deep}}: 
	\begin{equation}
	    f(\mathbf{X}) = \phi\left(\sum_{u \in \mathcal{V}}^{} \psi(\mathbf{x}_u)\right). \label{eqn:perm}
	\end{equation}
    As a result of the constraint, a permutation invariant function will produce the same output regardless of the order of the input nodes. These functions provide a summary or ``global" output for an input graph; however, ``local" relationships are usually the primary focus of interest \mbox{\cite{bronstein2021geometric}}. A permutation-invariant function can be applied locally, resulting in a permutation-equivariant operation (\textit{i.e.}, computing a certain permutation of inputs will result in the same output regardless of the order of those nodes) \mbox{\cite{bronstein2021geometric}}.
    
    For the proposed histogram layer, there are several useful properties that are inherited from this architecture selection. First, the standard histogram operator is permutation and rotationally invariant. In reference to Equation {\ref{eqn:perm}}, a standard histogram operation is computed for a single scalar feature $x$ where $\psi$ is a vector-valued function whose output dimension is given by the number of bins $B$. If $x$ falls in bin $j$, then $\psi_j(x) = 1$ and $0$ for all the other bins. To normalize the aggregated bin counts, we select $\phi$ to be a simple average function (\textit{i.e.}, divide by the number of elements). This selection of $\phi$ follows \mbox{\cite{bronstein2021geometric}}, which defines the identity and average functions to create linear-equivariant operators. Following the argument presented above, if the histogram operator is globally permutation and rotationally invariant, then a normalized histogram will result in permutation and rotational-equivariance. To further enhance the utility of the proposed histogram layer, a local sliding window operation will inherit the translational-equivariance property of the convolution operation.
    
    The concept of permutation invariance and equivariance can be generalized from sets to graphs \mbox{\cite{bronstein2021geometric}}. The permutation, rotational, and translational equivariance easily extends to images. As a result, the proposed histogram layer has desirable properties to successfully describe the notion of texture in local regions. In this work, we primarily focus on the utility of the layer for image data, but the properties of the histogram layer could be easily extended to non-visual modalities such as time signals and graphs. 
	
	\subsection{Previous Histogram Layers}
	The standard histogram operation counts the number of values that fall within certain ranges. The center of these ranges are defined as ``bin centers" and the interval or size of each range (or ``bin") is defined by the ``bin width." The standard histogram operation can be formulated as an indicator function. Consequently, this standard histogram operation cannot be easily used in ANNs because the functional form does not have the operation in terms of the histogram parameters (\textit{i.e.}, the bin centers and widths). Additionally, a standard histogram operation is not differentiable and,  thus, cannot be directly updated via backpropagation \cite{wang2016learnable}. In order to overcome the shortcomings of the standard histogram operation, two histogram layers were proposed for applications other than texture analysis. The first histogram layer was developed for semantic segmentation and object detection \cite{wang2016learnable} by Wang et al. The histogram operation was completed using a linear basis function to backpropagate the error to learn bin centers and widths. Wang et al.'s histogram layer has a convenient property in that it is implemented using pre-existing layers. The second histogram layer was developed for steganalysis \cite{sedighi2017histogram} and the histograms were modeled using RBFs. Sedighi and Fridich did not update the bin centers and widths, but these values were fixed to simulate the submodels of the projection spatial rich model (PSRM) \cite{holub2013random}.
	
	The histogram layer proposed in this work inherits properties from each of these models, but also incorporates novel aspects for texture analysis. The histogram layer will use RBFs to represent the histogram structure and this will provide smooth functions to update the bin centers and widths of the model. There are three key differences between our histogram layer and its predecessors. 1) Each of the previous approaches constructed global histograms. Spatial relationships are important in applications involving texture \cite{ojala2004texture,srinivasan2008statistical} as shown in Figure \ref{fig:Toy_example} and a localized approach will retain this information. 2) We investigate different bin numbers for the histogram layer as opposed to the previous methods that only used a single bin number as their hyperparameter. 3) The histogram layer can be placed anywhere in a network.

\section{Proposed Histogram Layer}
	\subsection{Soft Binning Operation}
\begin{figure}[htb]
	    \centering
		\begin{subfigure}{.45\textwidth}{
		\centering
	    \includegraphics[draft=false,width=1\linewidth]{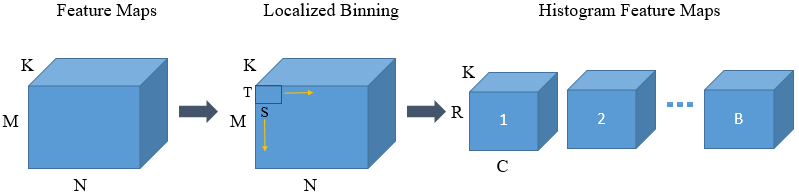}
				\caption{Visualization of localized histogram operation.}
				\label{fig:ex_binning}
			}
		\end{subfigure}
		\centering
		\begin{subfigure}{.45\textwidth}{
		\centering	\includegraphics[draft=false,width=1\linewidth]{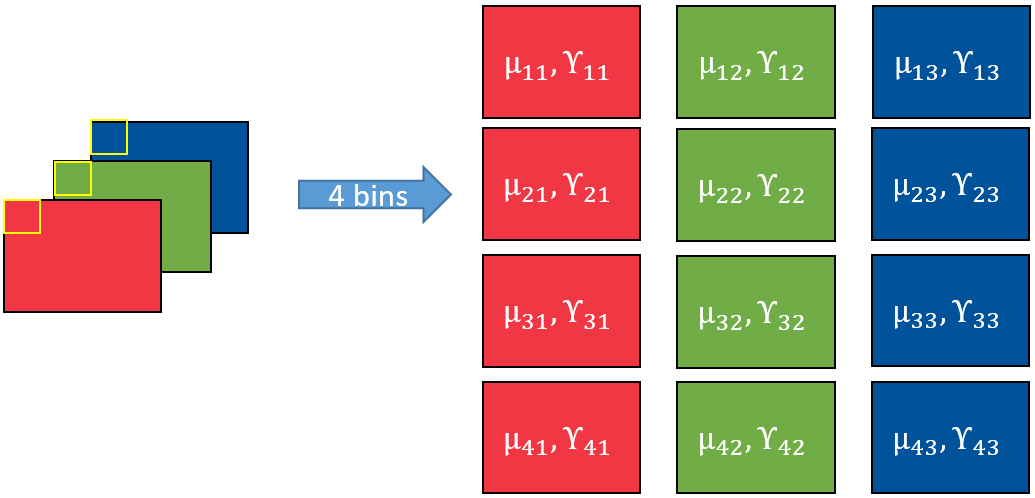}
				\caption{Example of binning process with 3-channel input (RGB).}
				\label{fig:hist_rgb_ex}
			}
		\end{subfigure}
	\caption{Visualization of localized histogram operation is shown in Figure \ref{fig:ex_binning}. For the histogram layer, the input is $K$ feature maps with spatial dimensions of $M \times N$. The normalized frequency count, $Y_{rcbk}$, can be computed with a sliding window of size $S \times T$ resulting in $B$ histogram feature maps of size $R \times C \times K$ where $B$ corresponds to the number of bins, $R$, $C$, and $K$ are the resulting output dimensions after binning the feature maps. In Figure \ref{fig:hist_rgb_ex}, an example 3-channel input is passed into a histogram layer with 4 bins. The output of the histogram layer is 12 histogram feature maps, where each feature map is a binned output with a distinct bin center ($\mu_{bk}$) and width ($\gamma_{bk}$) for each $k^{th}$ input channel.}
	\label{fig:Histogram_Operation}
	\end{figure}

	 We model our histogram with RBFs as done in Sedighi and Fridich's histogram layer \mbox{\cite{sedighi2017histogram}}. Unlike their implementation, we learn multiple bin widths to allow more model flexibility to estimate the distribution of features in the network. RBFs provide smoother approximations for histograms and RBFs have a maximum value of 1 when the feature value is equal to the bin center and the minimum value approaches 0 as the feature value moves further from the bin center. Also, RBFs are more robust to small changes
	 in bin centers and widths than the standard histogram operation because there is some allowance of error due to the soft binning assignments and the smoothness of the RBF. The means of the RBFs ($\mu_{bk}$) serve as the location of each bin (\textit{i.e.}, bin centers), while the bandwidth ($\gamma_{bk}$) controls the spread of each bin (\textit{i.e.}, bin widths), where $b$ is the number of histogram bins and $k$ is the feature channel in the input feature tensor. The normalized frequency count, $Y_{rcbk}$, is computed with a sliding window of size $S \times T$, and the binning operation for a histogram value in the $k^{th}$ channel of the input $x$ is defined as: 
	\begin{equation}
	Y_{rcbk} =  
	\cfrac{1}{ST}\sum_{s=1}^{S}\sum_{t=1}^{T}e^{-\gamma_{bk}^2\left(x_{r+s,c+t,k}-\mu_{bk}\right)^2}
	\label{equation:RBF}
	\end{equation}
	where $r$ and $c$ are spatial dimensions of the histogram feature maps.
	The process of aggregating the feature maps is shown in Figure \ref{fig:Histogram_Operation}.
	
\begin{figure*}[htb]
	\begin{center}
		\includegraphics[draft=false,width=.92\linewidth]{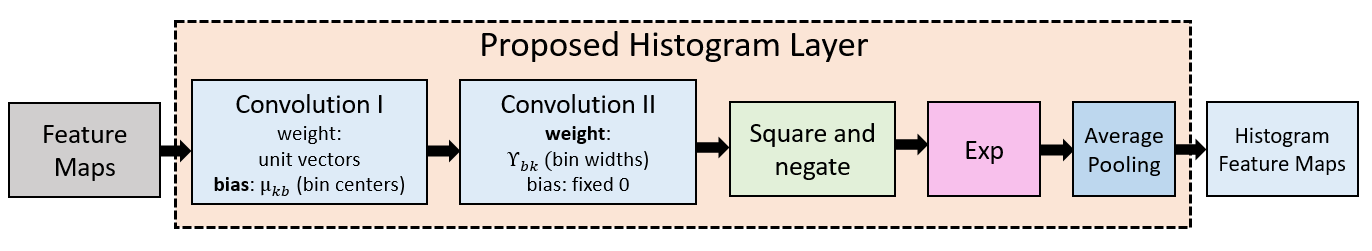}
	\end{center}
	\caption{Histogram layer implementation using pre-existing layers. The figure is adapted from \mbox{\cite{wang2016learnable} for comparison.}}
	\label{fig:Histogram_implementation}
\end{figure*}
	
	\subsection{Backpropagation} The histogram layer supports end-to-end learning through backpropagation to update the bin centers and widths as shown in \cite{wang2016learnable}. Each $k^{th}$ channel of the input $x$ is binned by the histogram in local spatial regions and stored in the $r^{th}$ row and $c^{th}$ column of the output of the histogram layer, $Y_{rcbk}$. The gradients for the parameters of the histogram layer with a window size of $S \times T$ are computed by Equations \ref{equation:grad_b} and \ref{equation:grad_w}:
	\begin{multline}
	\cfrac{\partial Y_{rcbk}}{\partial \mu_{bk}} = \cfrac{2}{ST} \sum_{s=1}^{S}\sum_{t=1}^{T}e^{-\gamma_{bk}^2\left(x_{r+s,c+t,k}-\mu_{bk}\right)^2} \times \\ \gamma_{bk}^2\left(x_{r+s,c+t,k}-\mu_{bk}\right)
	\label{equation:grad_b}
	\end{multline}
	\begin{multline}
	\cfrac{\partial Y_{rcbk}}{\partial \gamma_{bk}} = \cfrac{-2}{ST} \sum_{s=1}^{S}\sum_{t=1}^{T}e^{-\gamma_{bk}^2\left(x_{r+s,c+t,k}-\mu_{bk}\right)^2} \times \\ \gamma_{bk}\left(x_{r+s,c+t,k}-\mu_{bk}\right)^2
	\label{equation:grad_w}
	\end{multline}
	where $\cfrac{\partial Y_{rcbk}}{\partial \mu_{bk}}$ and $\cfrac{\partial Y_{rcbk}}{\partial \gamma_{bk}}$ are partial derivatives of $Y_{rcbk}$ with respect to the bin centers and widths of the histogram layer. In \cite{sedighi2017histogram}, the tails of the RBFs were set to 1 resulting in the gradient becoming zero if the feature map value is outside of every bin center's range. In our histogram layer, the gradients are a function of the distance between the feature map value and the bin centers. The gradient contribution from a feature map value will be small if it is far away from every bin center (\textit{i.e.}, outliers).
	
	\subsection{Implementation}
	The histogram layer is implemented using commonly used pre-existing layers as shown in Figure \ref{fig:Histogram_implementation}. As done in \cite{zhang2017deep,xue2018deep}, a $1 \times 1 \times K$ convolution is used to reduce the number of input feature maps, where $K$ is the new dimensionality of the feature maps. After the dimensionality reduction, the binning process starts by first assigning each feature value to every bin center (subtracting $\mu_{bk}$). The centering of the features to each bin is calculated by applying a $1 \times 1 \times B$ convolution to each feature map. The weights in the convolution kernels are fixed to 1 and each bias serves as the learnable bin center. 
	
	After the features are assigned to the bins, the centered features are then multiplied by the bandwidth ($\gamma_{bk}$) to incorporate the spread of the features for each bin. The incorporation of the spread of each bin is also computed by applying a $1 \times 1 \times B$ convolution to each feature map with the weights serving as the learnable bin widths and fixing the biases to be 0.  The contribution to each bin is calculated through RBF activation functions in Equation \ref{equation:RBF} by squaring, negating, and applying the exponential to the centered and scaled features. The contribution of each feature to every bin is between 0 and 1. The contributions of features in local spatial regions are then counted through average pooling (window size of $S \times T$ to produce feature maps of spatial dimensions $R \times C$) to compute the normalized frequency count of features belonging to each bin.

\section{Experimental Procedure}
	\subsection{Datasets} \label{sect:datasets}
\begin{figure}[t]
    \centering
	\begin{subfigure}{.15\textwidth}{
			\includegraphics[width=\textwidth]{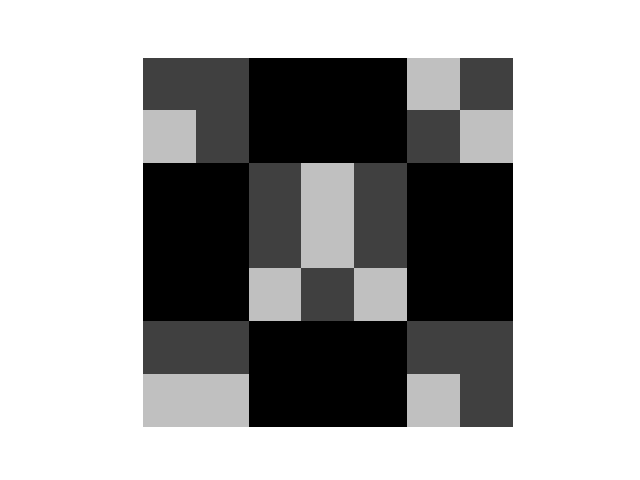}
			\caption{Checkerboard 1}
			\label{fig:dist_1A}
		}
	\end{subfigure}
   \centering
	\begin{subfigure}{.15\textwidth}{
			\includegraphics[width=\textwidth]{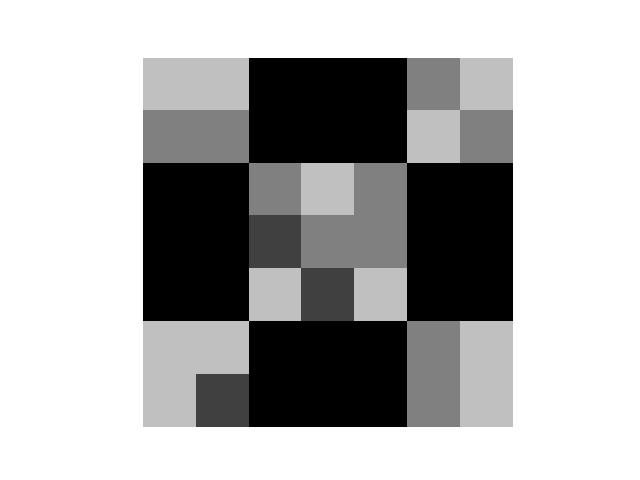}
			\caption{Checkerboard 2}
			\label{fig:dist_1B}
		}
	\end{subfigure}
   \centering
	\begin{subfigure}{.15\textwidth}{
			\includegraphics[width=\textwidth]{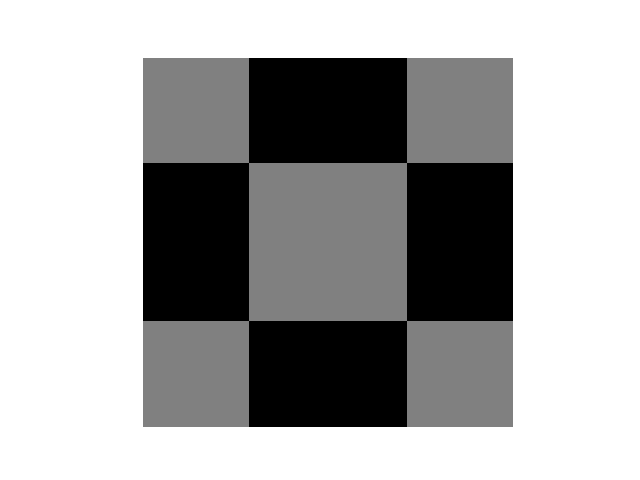}
			\caption{Checkerboard 3}
			\label{fig:dist_1C}
		}
	\end{subfigure}
    \centering
	\begin{subfigure}{.15\textwidth}{
			\includegraphics[width=\textwidth]{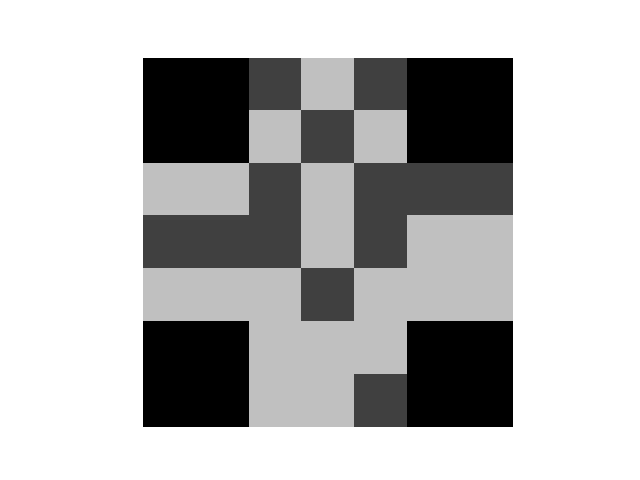}
			\caption{Cross 1}
			\label{fig:dist_2A}
		}
	\end{subfigure}
   \centering
	\begin{subfigure}{.15\textwidth}{
			\includegraphics[width=\textwidth]{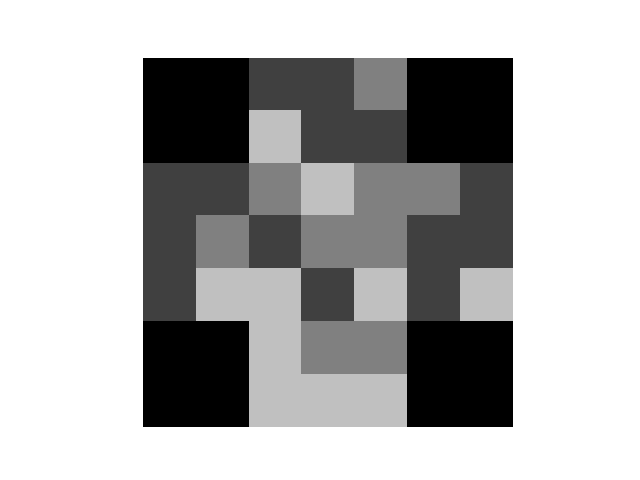}
			\caption{Cross 2}
			\label{fig:dist_2B}
		}
	\end{subfigure}
   \centering
	\begin{subfigure}{.15\textwidth}{
			\includegraphics[width=\textwidth]{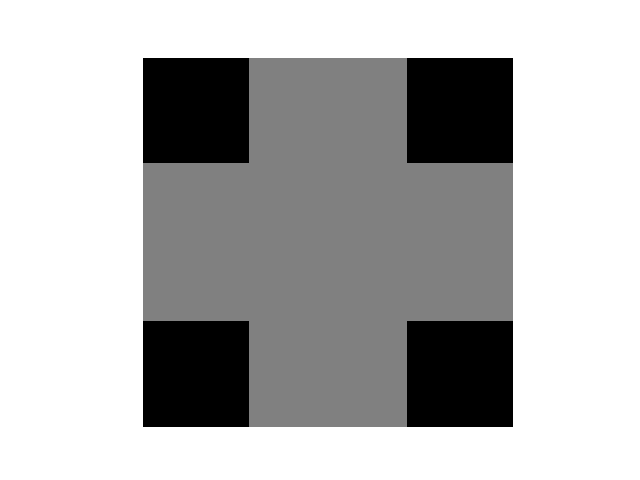}
			\caption{Cross 3}
			\label{fig:dist_2C}
		}
	\end{subfigure}
    \centering
	\begin{subfigure}{.15\textwidth}{
			\includegraphics[width=\textwidth]{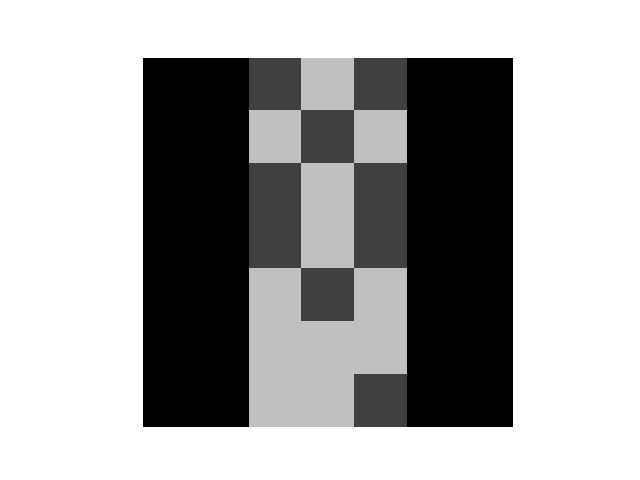}
			\caption{Stripe 1}
			\label{fig:dist_3A}
		}
	\end{subfigure}
   \centering
	\begin{subfigure}{.15\textwidth}{
			\includegraphics[width=\textwidth]{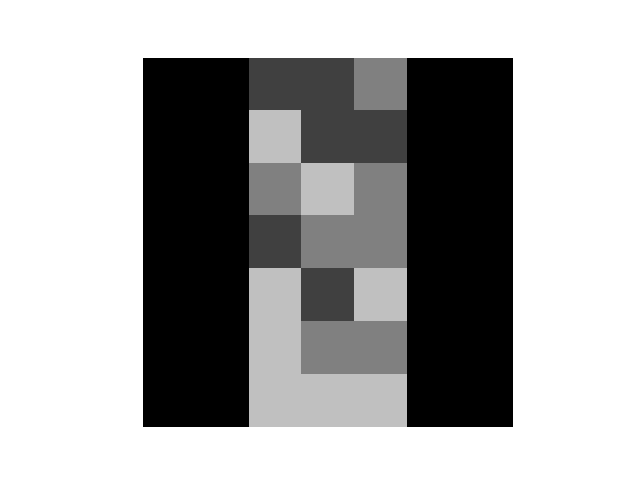}
			\caption{Stripe 2}
			\label{fig:dist_3B}
		}
	\end{subfigure}
   \centering
	\begin{subfigure}{.15\textwidth}{
			\includegraphics[width=\textwidth]{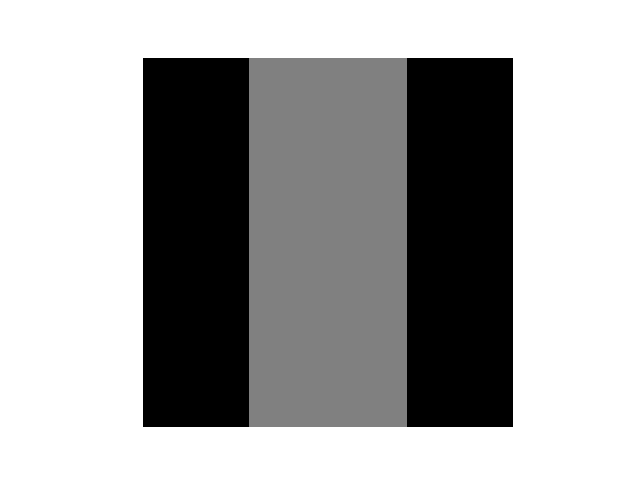}
			\caption{Stripe 3}
			\label{fig:dist_3C}
		}
	\end{subfigure}
	\caption{Example images of structural and statistical textures for the synthetic dataset were generated from sampling different distributions as described in Section \mbox{\ref{sect:datasets}}. The columns above show different structural textures while the rows correspond to different statistical textures. A total of three different structural and statistical textures comprised the dataset of nine distinct classes.}
	\label{fig:Toy_Set} 
\end{figure}

	For the synthetic dataset experiments, we generate 900 grayscale images of three distinct a) structural and b) statistical textures for a total of 9 classes (100 images per class). The structural textures were 1) cross, 2) checkerboard, and 3) stripes while the statistical labels were generated from a sampling of three distinct distributions (see Figure \mbox{\ref{fig:Toy_Set}}). For the first statistical class, a binomial distribution ($p = .5$) was used where either an intensity value of $64$ or $192$ was selected for each pixel. The second statistical class was sampled from a multinomial distribution with the following three intensity value choices which were of equal probability ($p = 1/3$): $64$, $128$, and $192$. The last statistical class only contained pixels with an intensity values of $128$. The distributions were selected so that the mean of the sampled pixels was approximately the same value (i.e., centered around an intensity values of 128). We then set the pixel values that do not overlap with the structure of interest to zero to generate the statistical and structural texture images. Through this experimental design choice, we require that the full distribution will be needed to learn to classify the different statistical labels of the data.
	
	Three material/texture datasets were investigated: Describable Texture Dataset (DTD) \cite{cimpoi14describing}, a subset of Materials in Context (MINC-2500) \cite{bell2015material}, and an extension of the ground terrain in outdoor scenes (GTOS-mobile) \cite{xue2018deep}. GTOS-mobile contains images for different resolutions (256$\times$256, 384$\times$384, and 512$\times$512) but only single-scale experiments using 256$\times$256 images were performed in this work. The total number of training and testing images for DTD, MINC-2500, and GTOS-mobile were 5,640, 54,625, and 37,381 respectively. The number of classes for DTD, MINC-2500, and GTOS-mobile were 47, 23, and 31 respectively. For DTD and MINC-2500, the published training and testing splits were used in each experiment (five and ten folds for DTD and MINC-2500 respectively). The ResNet50 architecture was used as the baseline model for these DTD and MINC-2500 datasets while ResNet18 was used for GTOS-mobile \cite{xue2018deep}. GTOS-mobile only has a single training and test split, but five experimental runs of different initialization were performed to investigate the stability of our model. 
	\subsection{Experimental Setup}

		\begin{figure*}[t]
		\begin{center}
			\includegraphics[draft=false,width=.80\linewidth]{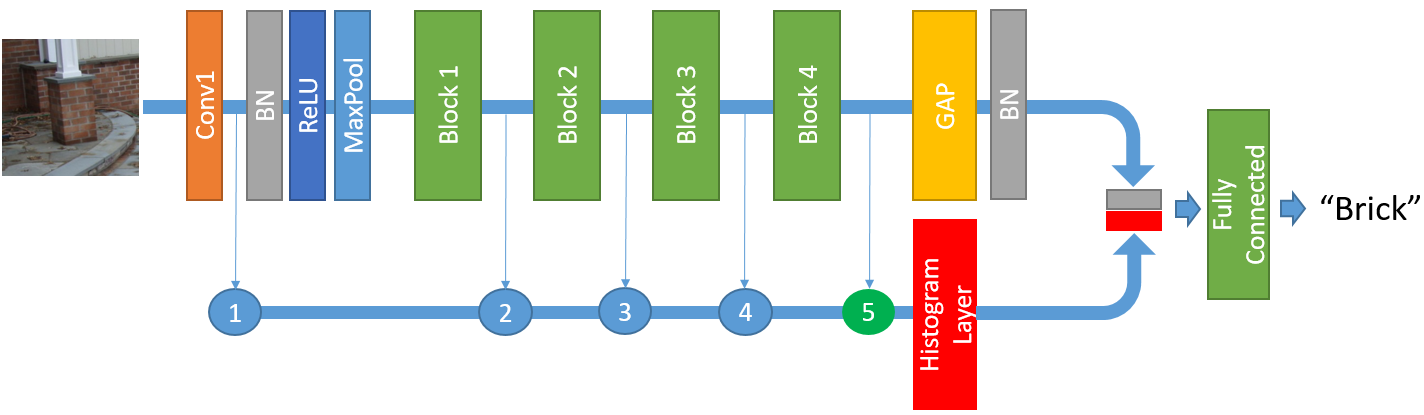}
		\end{center}
		\caption{Histogram Layer for ResNet with $B$ bins (HistRes\_$B$) based on ResNet18 and ResNet50 \cite{he2016deep}. Each block is comprised of convolutional, max pooling, and batch normalization layers. The convolutional features from the model are passed into the global average pooling (GAP) and the histogram layer to capture texture, spatial, and orderless convolutional features. The features are then concatenated together before being fed into a fully connected layer for classification. The location of the histogram layer was varied from 1 to 5. In this figure, the feature maps from the last convolution layer (location 5) are passed into the histogram layer.}
		\label{fig:HistNet}
	\end{figure*}
	In order to provide more insight into the histogram layer, we performed experiments with a synthetic dataset. Unlike standard convolution layers, the histogram layer has a unique property to directly model statistical texture. As noted in recent work {\cite{zhu2021learning}}, convolution layers can capture structural textures as opposed to statistical textures. The ability to model statistical textures is crucial in several texture analysis tasks \mbox{\cite{haralick1973textural,liu2019texture,humeau2019texture}}. In addition to investigating the notion of statistical versus structural textures, we also analyzed the performance of global versus local feature aggregation as well as analyzing the combination of statistical and structural features.
	
	In addition to the synthetic dataset experiments, we performed material/texture classification and compared  the proposed histogram layer to other texture encoding approaches. The experiments consisted of two components: 1) configuration of histogram layer and 2) comparison to other texture encoding approaches. For the first component, we investigated the effects of a) normalization, b) sum-to-one constraint across bins, and c) location of histogram layer. Previous works incorporated batch normalization of their convolutional features before fusing their texture features (\textit{i.e.}, concatenation, weighting) \cite{hu2019multi,xue2018deep,zhang2017deep}. We wanted to investigate whether or not the impact of batch normalization for our new layer would improve results similar to existing ``encoding layers." In normalized histograms, the heights of the bins (\textit{i.e.}, counts within each bin) are constrained to sum up to one across all bins. We also wanted to compare if including or relaxing this constraint would affect overall performance and/or increase the robustness of the histogram layer. The final part of the configuration of the histogram layer was analysis of the effect of location in the networks. The histogram layer was added at the end of the network for the normalization and sum-to-one constraint experiments, but it may be more advantageous to learn texture information earlier in the network. Previous histogram-based features used lower level features (\textit{e.g.}, edges \cite{frigui2008detection} for edge histogram descriptors); also, and learning histograms of convolutional features earlier in the network may exploit more texture information.
	
	\subsection{Training Details}
	The synthetic dataset was divided into 70/10/20 split for training, validation, and test sets. The image sizes were set to 3$\times$3 and 7$\times$7 for the local and global experiments respectively. Each model was trained for 300 epochs and early stopping was used to terminate training if the validation loss did not decrease within 10 epochs. To train the model, we used a batch size of 64, cross entropy loss, Adam optimization with a maximum learning rate of $.001$, and momentum parameters of $\beta_{1}= .9$ and $\beta_{2}= .999$. 
	
	A similar training procedure from \cite{hu2019multi,xue2018deep} was used for the texture/material classification experiments. For each dataset, all the images were first resized to 256$\times$256, and a random crop of 80 to 100\% of each image was extracted with a random aspect ratio of 3/4 to 4/3. The cropped images were then resized to 224$\times$224 and normalized by subtracting the per-channel mean and dividing by the per-channel standard deviation. Random horizontal flip ($p=.5$) was also used for data augmentation. The training settings for each network were the following: batch size of 128 (64 for the DTD and MINC-2500 models in the scale experiments), cross-entropy loss function, SGD with momentum ($\alpha=.9$), learning rates decay every 10 epochs by a factor of .1 and training was stopped after 30 epochs. The initial learning rates for the newly added and pre-trained layers were .01 and .001 respectively.

	\subsection{Architectures and Evaluation Metrics}
	\label{sect:architecture} 
	A simple model was used for the synthetic dataset experiments. The network consisted of a feature extractor (\textit{i.e.}, histogram or convolutional layer), GAP layer, and fully connected output layer. The filter sizes for the histogram and convolution layer were set to be 3 $\times$ 3. We evaluated each model as a global and local feature extraction approach by applying the model to two different image sizes (3$\times$3 and 7$\times$7). Using three random seeds, we set each model to the same random initialization to fairly compare each approach and evaluate the discriminative power of the different features. We also set the number of output feature maps for each method to be equal to the number of unique intensity values (three). The histogram layer model was initialized to have equally spaced bins and equal widths that covered the range of the input images, $[0,1]$, as in the case of a standard histogram. In addition to the two separate feature extraction approaches, we also ran a combination of the histogram and convolutional features by concatenating output feature vectors before the output layer. The number of parameters of the convolutional model was 66, while the histogram model had 42 parameters, and the combination models had 99 parameters.
	
	Two pre-trained ResNet models (\textit{i.e.}, ResNet18 and ResNet50) were used as the baseline for the convolutional features for the texture/material classification experiments. The models that incorporated the histogram layer are referred to as HistRes\_$B$, where $B$ is the number of bins. The HistRes\_$B$ architecture is shown in Figure \ref{fig:HistNet}. For the normalization and sum-to-one constraint experiments, the number of bins was also varied, in order to investigate the effects of adding additional histogram feature maps to the network. After the normalization and sum-to-one constraint experiments were completed, we investigated the histogram layer in each model. The kernel size and stride for the histogram was selected to produce local feature maps with $R = 2$ and $C = 2$ for each location; in this manner, the number of histogram features were equal to the number of convolutional features from the global average pooling (GAP) layer (512 and 2048 for ResNet18 and ResNet50 respectively). 
	
	For the channel-wise pooling, the value of $K$ was selected so that the number of features from the GAP layer and the histogram layer were equal in order to leverage the contribution of texture, spatial, and orderless convolutional features. For Resnet18, the input number of feature maps from each scale was reduced to 32, 16, and 8 for HistRes\_4, HistRes\_8, and HistRes\_16 respectively. For Resnet50, the input number of feature maps from each scale was reduced to 128, 64, and 32 for HistRes\_4, HistRes\_8, and HistRes\_16 respectively. The bin centers and widths were initialized to values sampled uniformly between $\left(\frac{-1}{\sqrt{BK}},\frac{1}{\sqrt{BK}}\right)$. After the best configuration of the HistRes model was determined, we compared the performance using overall accuracy between different ``encoding layers" for each version of ResNet: 1) global average pooling (GAP); 2) Deep Texture Encoding Network (DeepTEN) \cite{zhang2017deep}; 3) Deep Encoding Pooling (DEP) \cite{xue2018deep}; 4) Fisher Vector CNN (FV-CNN) \cite{cimpoi2015deep}; 5) Multi-level texture encoding and representation (MuLTER) \cite{hu2019multi}; and 6) our HistRes\_$B$. We also implemented the proposed model using the piecewise linear basis function to analyze differences between the two binning approximations. Our implementation of the linear basis function differs from the original work because the histogram layer is applied directly to the convolutional feature maps and not to the class likelihood maps \cite{wang2016learnable}. 

	\section{Results}
	\begin{figure*}[htb!] \label{fig:toy_example}
    \centering 
		\begin{subfigure}{.24\textwidth}{
				\includegraphics[draft=false,width=\textwidth]{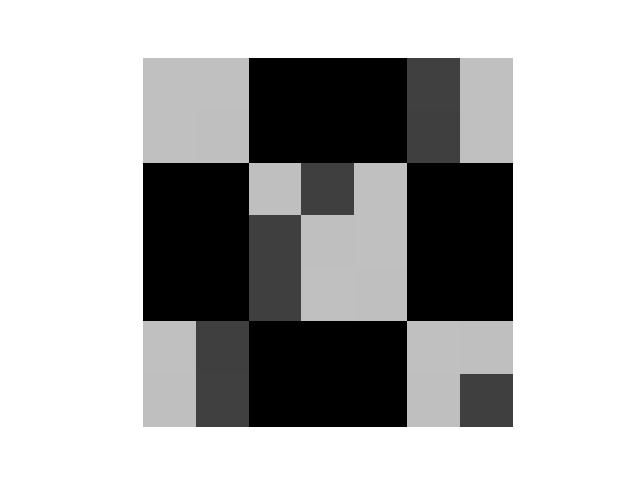}
				\caption{Input Image}
				\label{fig:image}
			}
		\end{subfigure}
		\begin{subfigure}{.36\textwidth}{
				\includegraphics[draft=false,width=\textwidth]{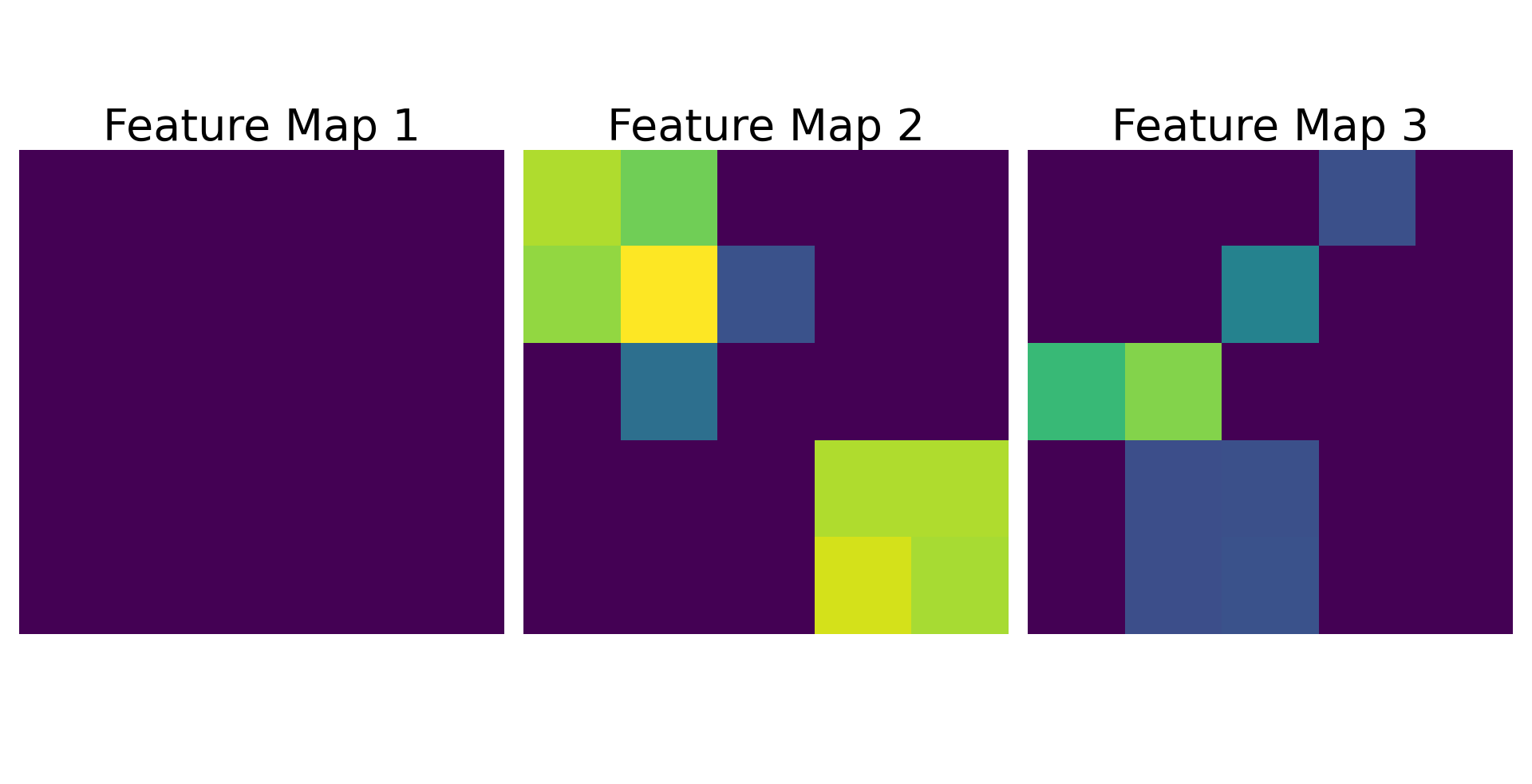}
				\caption{Convolutional Feature Maps}
				\label{fig:CNN}
			}
		\end{subfigure}
		\begin{subfigure}{.36\textwidth}{
				\includegraphics[draft=false,width=\textwidth]{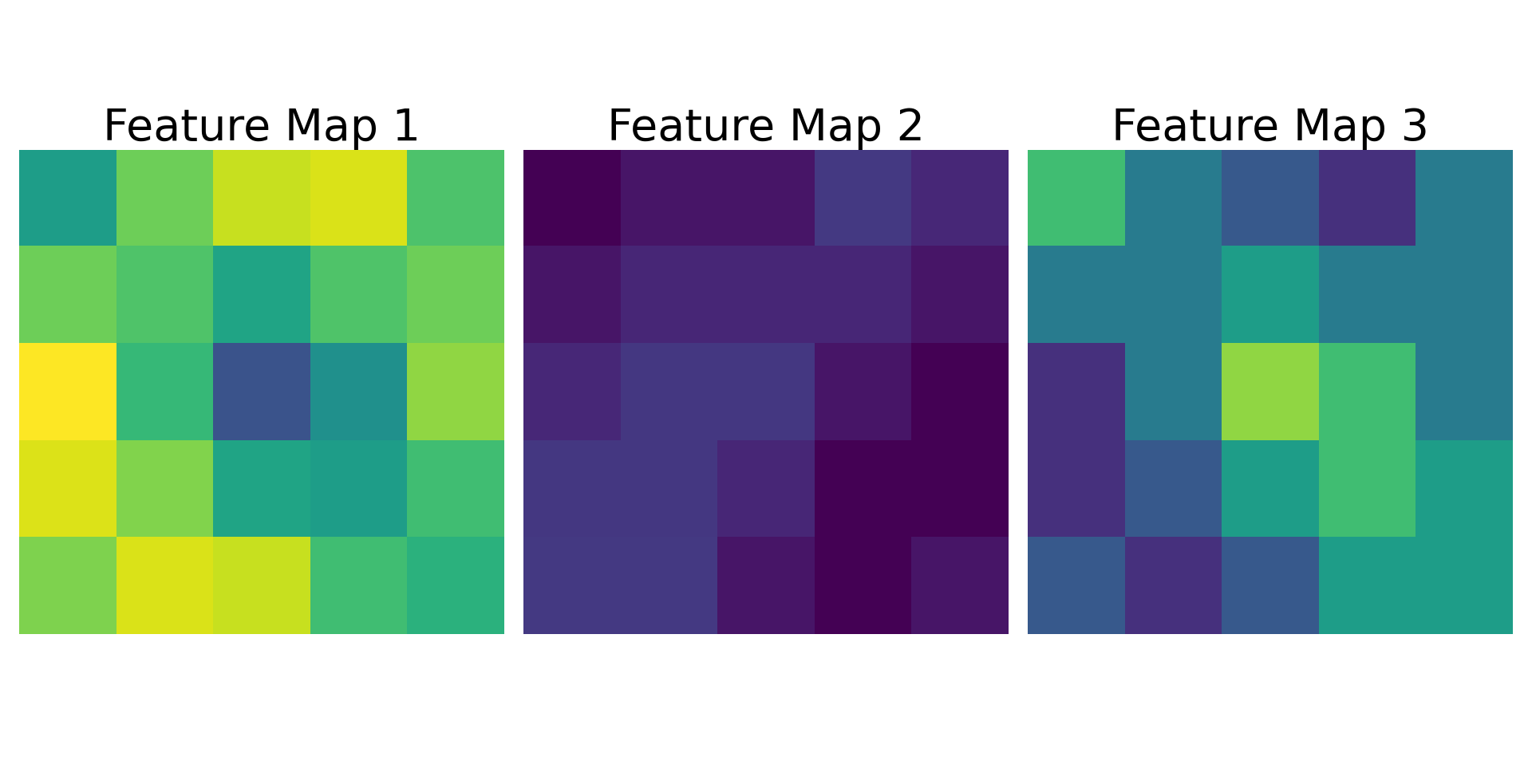}
				\caption{Histogram Feature Maps}
				\label{fig:Hist}
			}
		\end{subfigure}
	\caption{Example outputs of convolutional and histogram features maps for an input image. The input image size was 7$\times$7 and the kernel size for both models was 3$\times$3. The convolutional model uses the standard ReLU activation function. As seen in this example, one feature map does not provide any information since all the negative values are set to 0 after the activation function. The histogram layer provides informative information based on the location of each pixel to the learned bin center and width. The bin centers of the histogram covered the low, mid, and high intensity values (feature maps 1 through 3). For feature map 1 in Figure {\ref{fig:Hist}}, the bin responses are higher for neighborhoods of lower intensity pixel values. With the histogram layer, we can infer explainable information as opposed to information from the convolutional feature maps.}
		\centering \label{fig:ex_output}
		\end{figure*}
	\subsection{Synthetic Statistical and Structural Textures}
		\begin{table}[htb]
		\caption{Test accuracy for each model \mbox{\ref{tab:Syn_A}}) globally and \mbox{\ref{tab:Syn_B}}) locally for convolutional and histogram layer features. The result with the best average across the local and global experiments is bolded.}
		\begin{subtable}[h]{.5\textwidth}
			\centering
			\begin{tabular}{|c|c|c|c|}
				\hline
				Label     & Convolution & Histogram  & Combination \\ \hline
				Both         & 34.4$\pm$16.5\% & 52.59$\pm$4.95\% & \textbf{90.67$\pm$2.27\%}          \\ \hline
				Statistical   & 41.8$\pm$6.10\%          & 90.22$\pm$1.26\% & \textbf{90.67$\pm$2.27\%} \\ \hline
				Structural & 77.8$\pm$31.4\%          & 58.81$\pm$4.92\% & \textbf{100.0$\pm$0.00\%} \\ \hline
			\end{tabular}
			\caption{Global}
			\label{tab:Syn_A}
		\end{subtable}
		\begin{subtable}[htb]{.5\textwidth}
			\centering
			\begin{tabular}{|c|c|c|c|}
				\hline
				Label   & Convolution & Histogram  & Combination \\ \hline
				Both         & 67.56$\pm$11.3\% & 80.14$\pm$6.34\% & \textbf{97.04$\pm$1.47\%}  \\ \hline
				Statistical   & 68.59$\pm$9.87\% & 92.30$\pm$4.54\% & \textbf{97.78$\pm$1.09\%}  \\ \hline
				Structural & 96.00$\pm$4.09\% & 85.18$\pm$11.3\% & \textbf{99.26$\pm$0.76\% } \\ \hline
			\end{tabular}
			\caption{Local}
			\label{tab:Syn_B}
		\end{subtable}
		\label{tab:Syn_dataset}
	\end{table}
	
The synthetic dataset experimental results are shown in Table {\ref{tab:Syn_dataset}}. We observe that the histogram model performed better overall on the test set in comparison the baseline convolutional model for both global and local experiments. For the global experiments, the convolutional model particularly struggled with the statistical distributions. These results support the hypothesis that convolutional models are biased towards identifying structural textures. The convolution operation is essentially a weighted average operator when the image size is equal to the kernel size (as is the case for the global experiments). Consequently, the discriminative power of the convolutional model cannot successfully distinguish the differences in the statistical textures. The histogram layer is better able than the convolutional layer to model a distribution that captures distinct information and distinguishes the statistical labels as opposed to the convolutional model. Since the histogram layer is used as a global feature extractor, the histogram model struggled with distinguishing the structural textures. The reason for this is due to the checkerboard and cross structures. The spatial arrangement of theses structures is different, but distribution of pixel values will be very similar in some instances. Average pooling is used to aggregate the ``binned" feature values and cannot account for the structural changes. This result is intuitive since the standard histogram operation is not sensitive to the ordering of samples.

The local experiments improved both the convolutional and the histogram models which further supports the conjecture that spatial information is important for texture analysis. However, the convolutional model is less able to distinguish the statistical information than is the histogram layer. The addition of spatial information did significantly improve the structural performance for the histogram layer and, not only can the histogram layer better classify the different texture classes, but we also gain insight from the feature maps of the network. In Figure {\ref{fig:ex_output}}, we show an example image of the checkerboard structure and pixels from the first statistical class. We can infer information from the input and output features maps based on the output ``counts" from the histogram layer ({\textit{i.e.}}, pixels in a neighborhood have similar values). In addition to seeing improvement with spatial information, we saw that the combination of convolutional and histogram features maximized the performance for both statistical and structural textures. As a result, jointly optimizing both convolutional and histogram layers have the potentially to maximize texture performance. We investigate this further on real texture/material datasets. 
	\subsection{Configuration of histogram layer}
	\subsubsection{Normalization} The first step of determining the best configuration of the histogram models was to investigate the effects of normalization. For the convolutional and histogram features, batch normalization and normalizing the count (averaging feature sum by window size) is used respectively. From Table \ref{tab:Normalization}, normalization of the features improved performance for each dataset, with the largest improvements occurring for the DTD dataset. The images in DTD are collected ``in-the-wild" (\textit{i.e.,} collected in a non-controlled environment); therefore, normalization plays an important role in assisting the model to account for large intra- and inter-class variations. For MINC-2500 and GTOS-mobile, the improvement did occur but only to a lesser extent. Both datasets are also collected ``in-the-wild," but each dataset is significantly larger than DTD and each also has a smaller number of classes, so normalization may not have led to comparably large improvements such as those shown by the models trained with DTD. Another finding was that the number of bins did not affect performance for either the normalized or unnormalized models. This shows that, unlike results from the conventional histogram-based feature approaches, the histogram layer performance does not rely heavily on the number of bins selected.
		\begin{table}[t]
		\caption{Test accuracy for each HistRes model \ref{tab:Norm_A}) with and \ref{tab:Norm_B}) without normalization for convolutional and histogram layer features. The result with the best average is bolded.}
		\begin{subtable}[h]{.5\textwidth}
			\centering
			\begin{tabular}{|c|c|c|c|}
				\hline
				Dataset     & HistRes\_4           & HistRes\_8  & HistRes\_16        \\ \hline
				DTD         & \textbf{71.98$\pm$1.23}\% & 71.62$\pm$0.80\% & 71.69$\pm$1.09\%          \\ \hline
				MINC-2500   & 82.14$\pm$0.31\%          & 81.97$\pm$0.47\% & \textbf{82.14$\pm$0.51\%} \\ \hline
				GTOS-mobile & 78.18$\pm$0.33\%          & 78.55$\pm$0.88\% & \textbf{78.56$\pm$0.71\%} \\ \hline
			\end{tabular}
			\caption{Batch normalization and normalized count}
			\label{tab:Norm_A}
		\end{subtable}
		\begin{subtable}[htb]{.5\textwidth}
			\centering
			\begin{tabular}{|c|c|c|c|}
				\hline
				Dataset     & HistRes\_4  & HistRes\_8  & HistRes\_16 \\ \hline
				DTD         & 62.58$\pm$3.27\% & 66.44$\pm$2.42\% & \textbf{68.07$\pm$1.86\%}  \\ \hline
				MINC-2500   & 80.03$\pm$1.23\% & 80.06$\pm$1.10\% & \textbf{81.38$\pm$1.08\%}  \\ \hline
				GTOS-mobile & 72.31$\pm$2.93\% & 73.82$\pm$2.08\% & \textbf{74.48$\pm$2.37\%}  \\ \hline
			\end{tabular}
			\caption{No batch normalization and unnormalized count}
			\label{tab:Norm_B}
		\end{subtable}
		\label{tab:Normalization}
	\end{table}
	
	\subsubsection{Sum-to-one}
	In addition to normalization, typical histogram bin votes are constrained to sum to one since a histogram estimates a probability density function (PDF) and a PDF integrates to one. In the normalization experiments, this constraint was relaxed. Ideally, relaxing this constraint will provide increased robustness to outliers. On the other hand, such action may also prevent learning if all the features initially start outside the range of the histogram (\textit{i.e.}, vanishing gradient will occur if features are in the tails of the RBFs). As shown in Table \ref{tab:Sum_constraint}, the enforcement of the constraint improved performance slightly for each dataset except for DTD. The model still retains robustness due to the soft binning assignments, but enforcing this constraint will assist in overcoming poor initialization by promoting each feature value to have some contribution to each bin. 
	\begin{table}[t]
	\caption{Test accuracy for each HistRes model \ref{tab:Sum_A}) with and \ref{tab:Sum_B}) without sum-to-one constraint across bins. The result with the best average is bolded.}
	\begin{subtable}{.5\textwidth}
		\centering
		\begin{tabular}{|c|c|c|c|}
			\hline
			Dataset     & HistRes\_4 & HistRes\_8          & HistRes\_16         \\ \hline
			DTD         & 71.62$\pm$1.05\% & 71.53$\pm$1.21\%          & \textbf{71.75$\pm$1.11\%}          \\ \hline
			MINC-2500   & 82.23$\pm$0.41\% & \textbf{82.31$\pm$0.44\%} & 82.42$\pm$0.33\%          \\ \hline
			GTOS-mobile & 78.64$\pm$0.76\% & 78.77$\pm$0.81\%          & \textbf{79.75$\pm$0.84\%} \\ \hline
		\end{tabular}
		\caption{Enforced sum-to-one constraint across bins}
		\label{tab:Sum_A}
	\end{subtable}
	\begin{subtable}[htb]{.5\textwidth}
		\centering
		\begin{tabular}{|c|c|c|c|}
			\hline
			Dataset     & HistRes\_4          & HistRes\_8 & HistRes\_16 \\ \hline
			DTD         & \textbf{71.98$\pm$1.23\%} & 71.62$\pm$0.80\% & 71.69$\pm$1.09\%  \\ \hline
			MINC-2500   & \textbf{82.14$\pm$0.31}\%          & 81.97$\pm$0.47\% & \textbf{82.14$\pm$0.51\%}  \\ \hline
			GTOS-mobile & 78.18$\pm$0.33\%          & 78.55$\pm$0.88\% & \textbf{78.56$\pm$0.71\%}  \\ \hline
		\end{tabular}
		\caption{Relaxed sum-to-one constraint across bins}
		\label{tab:Sum_B}
	\end{subtable}
	\label{tab:Sum_constraint}
	\end{table}
	\subsubsection{Location of histogram layer} 
	\begin{table*}[htb]
		\caption{Test accuracy for varying location of histogram layer for best histogram model for each dataset: HistRes\_$4$ with no sum-to-one constraint for DTD and HistRes \_$16$ with the sum-to-one constraint (all features for each model were normalized). The number of features for each location was constrained to be the same which lead to different window sizes and strides for each location. The kernel sizes for locations 1 through 5 were 64, 32, 16, 8 and 4 respectively. The strides for locations were 32, 16, 8, 4 and 2. The result with the best average is bolded.}
		\centering
		\begin{tabular}{|c|c|c|c|c|c|}
			\hline
			Dataset      & Location 1             & Location 2    & Location 3    & Location 4    & Location 5             \\ \hline
			DTD              & 71.28$\pm$1.03\%          & 71.37$\pm$0.99\% & 71.16$\pm$1.02\% & 71.13$\pm$1.00\% & \textbf{71.50$\pm$0.78\%} \\ \hline
			MINC-2500 & \textbf{83.07$\pm$0.54\%} & 82.96$\pm$0.57\% & 82.96$\pm$0.27\% & 83.04$\pm$0.71\% & \textbf{83.07$\pm$0.42\%} \\ \hline
			GTOS-mobile  & 78.63$\pm$1.00\%          & 78.57$\pm$0.61\% & 78.97$\pm$0.82\% & 79.12$\pm$1.70\% & \textbf{79.75$\pm$0.84\%} \\ \hline
		\end{tabular}
		\label{tab:Scale}
	\end{table*}
	
	\begin{table*}
		\centering
		\caption{Test accuracy of each encoding method. The results for DeepTEN \cite{zhang2017deep}, DEP \cite{xue2018deep}, and FV-CNN \cite{cimpoi2015deep} are reported from \cite{xue2018deep} while MuLTER is reported from \cite{hu2019multi}. The results we obtained running the baseline ResNet model with global average pooling (GAP) are indicated by GAP* while GAP was reported from \cite{xue2018deep}. For our experiments, we average our results for each data split and show a 1-standard deviation to show the stability of our method. We compare the best histogram model: HistRes\_$4$ for DTD and HistRes\_$16$ for MINC-2500 and GTOS-mobile. We also implemented the piecewise linear binning function \cite{wang2016learnable}
		(LinearHist\_$B$) to compare with our RBF model using the same number of bins.}
        \begin{tabular}{|c|c|c|c|c|c|c|c|c|}
            \hline
            Dataset     & GAP   & GAP*                    & DeepTEN & DEP           & FV-CNN & MuLTER & LinearHist\_B  & HistRes\_B              \\ \hline
            DTD         & -     & 73.07 $\pm$0.79\%         & 69.6\%    & \textbf{73.2\%} & 72.3\%   & -      & 71.53$\pm$0.80\% & 71.98$\pm$1.23\%          \\ \hline
            MINC-2500   & -     & \textbf{83.01$\pm$0.38\%} & 80.4\%    & 82.0\%          & 63.1\%   & 82.2\%   & 82.35$\pm$0.45\% & 82.42$\pm$0.33\%          \\ \hline
            GTOS-mobile & 70.82\% & 76.09$\pm$0.91\%          & 74.2\%    & 76.07\%         & -      & 78.2\%   & 78.70$\pm$0.57\% & \textbf{79.75$\pm$0.84\%} \\ \hline
        \end{tabular}
		\label{tab:test_acc}

	\end{table*}	
	For texture analysis, the location/scale at which features are extracted are important. Most texture ``encoding layers" are placed at the end of the network. Intuitively, this makes sense because CNNs learn simple features such as edges and corners early in the network and then progress towards domain-specific information. The features extracted at the end of the network represent larger areas of the image (\textit{i.e.}, larger scales). Since pre-trained models are used, ``encoding layers" are placed at the end to tailor these features towards the application of interest. We wanted to verify this hypothesis by placing the best histogram model for each dataset at various locations in the network. For DTD, we used HistRes\_$4$ with no sum-to-one constraint, while for both GTOS-mobile and MINC-2500 we used HistRes\_$16$ with the sum-to-one constraint enforced. 
	
	From Table \ref{tab:Scale}, it is apparent that performance did not vary much with the location of the histogram layer. This shows that the histogram layer can use both simple- and domain-specific features to perform texture classification. As a result, this new layer can be placed anywhere in the network and achieve the same level of performance by learning the distribution of low- to high-level features. The number of features for each location of the histogram was constrained to be the same ($512$ and $2048$ for ResNet18 and ResNet50 respectively). An interesting future work would be to remove this constraint and use the same window size for each location in the network to further investigate the impact of the location of the histogram layer for performance. 
	\subsection{Comparison to other texture encoding approaches}
    \begin{table*}[h]
	\caption{The log of the Fisher's Discriminant Ratio (FDR) \cite{bishop2006pattern} was computed on the features extracted from the 10,000 randomly sampled training images from the GTOS-mobile dataset (the same random images were used for each model). The FDR measures the ratio of the inter- (\textit{i.e.}, between class) and intra- (\textit{i.e.}, within class) class variance. Larger FDRs indicated that the samples within a class were close to each other and well-separated from other samples in different classes. The FDR scores for the features of our model were better. For 19 of the 31 classes, the log of the FDR was larger for the HistRes\_$16$ model. Though the FDRs for the training data were better for HistRes\_$16$ than GAP, our proposed model still generalized well to the test set in comparison to the baseline model (as indicated in Figure \ref{fig:CMTSNE_visual}). The best average for the log of the FDR for each class is bolded.}
	
		\begin{subtable}[htb]{.49\textwidth}
			\centering
        \begin{tabular}{|c|c|c|c|}
            \hline
            Class Name       & GAP  & LinearHist\_16 & HistRes\_$16$  \\ \hline
            Painting         & 37.42$\pm$1.44 & \textbf{38.40$\pm$1.13} & {38.08$\pm$1.08}               \\ \hline
            aluminum         & 39.77$\pm$2.51  & 39.62$\pm$1.53& \textbf{39.90$\pm$1.85}                \\ \hline
            asphalt          & 7.37$\pm$0.13 & \textbf{33.74$\pm$1.35}& {27.66$\pm$0.57}               \\ \hline
            brick            & 36.09$\pm$1.03 & 37.68$\pm$1.03 & \textbf{37.76$\pm$0.50}                \\ \hline
            cement           & 7.54$\pm$0.13 & \textbf{32.07$\pm$1.12} & {25.21$\pm$2.44}               \\ \hline
            cloth            & \textbf{39.95$\pm$0.94} &37.75$\pm$0.96 & 36.28$\pm$1.28                        \\ \hline
            dry\_leaf        & \textbf{40.19$\pm$3.09} &39.13$\pm$2.15 & 38.60$\pm$1.80                          \\ \hline
            glass            & 37.05$\pm$2.10  & \textbf{38.75$\pm$1.27} & {38.34$\pm$1.26}               \\ \hline
            grass            & 39.02$\pm$1.42  &38.85$\pm$2.67 & \textbf{39.92$\pm$1.38}               \\ \hline
            large\_limestone & 38.78$\pm$1.95  &37.95$\pm$1.84 & \textbf{39.14$\pm$1.15}               \\ \hline
            leaf             & \textbf{39.66$\pm$1.56} &38.75$\pm$0.63 & 38.77$\pm$1.77                        \\ \hline
            metal\_cover     & 37.64$\pm$1.79  &36.68$\pm$1.12 & \textbf{37.88$\pm$1.59}               \\ \hline
            moss             & \textbf{38.99$\pm$1.78} &37.96$\pm$0.99 & 38.97$\pm$2.15                        \\ \hline
            paint\_cover     & \textbf{41.43$\pm$0.41}  &40.36$\pm$1.11 & 40.39$\pm$1.82                        \\ \hline
            paint\_turf      & 38.57$\pm$0.79  &38.85$\pm$1.15    & \textbf{39.17$\pm$1.39}               \\ \hline
            paper            & \textbf{41.20$\pm$1.14}  &38.49$\pm$1.24 & 40.18$\pm$1.35                        \\ \hline
        \end{tabular}
		\caption{FDR Scores for first 16 classes}
		\label{tab:FDR_A}
		\end{subtable}
		\begin{subtable}[h]{.49\textwidth}
			\centering
        \begin{tabular}{|c|c|c|c|}
            \hline
            Class Name       & GAP      & LinearHist\_16 & HistRes\_$16$                \\ \hline
            pebble           & 41.78$\pm$1.71  & 39.70$\pm$0.85 & \textbf{42.57$\pm$1.50}             \\ \hline
            plastic          & 41.54$\pm$1.34    & 37.42$\pm$1.27          & \textbf{41.91$\pm$1.00}             \\ \hline
            plastic\_cover   & \textbf{39.30$\pm$1.27}    & 38.85$\pm$1.30   & 38.87$\pm$1.68                      \\ \hline
            root             & 37.58$\pm$1.07       & \textbf{37.81$\pm$1.28}    & {37.60$\pm$1.67}             \\ \hline
            sand             & {37.96$\pm$17.90}   & \textbf{39.43$\pm$1.01}    & 35.73$\pm$16.64                     \\ \hline
            sandPaper        & \textbf{42.84$\pm$1.81}   & 39.76$\pm$0.53  & 41.79$\pm$2.22                      \\ \hline
            shale            & 41.11$\pm$1.20     & 39.41$\pm$0.64 & \textbf{41.25$\pm$0.97}             \\ \hline
            small\_limestone & 39.17$\pm$1.06     & 36.72$\pm$0.81                          & \textbf{39.19$\pm$1.07}             \\ \hline
            soil             & \textbf{39.09$\pm$0.83}   & 36.25$\pm$2.08    & 38.92$\pm$0.74                      \\ \hline
            steel            & 39.71$\pm$1.61   & \textbf{42.55$\pm$2.01}                       & {40.30$\pm$1.03}              \\ \hline
            stone\_asphalt   & 42.91$\pm$0.90   & 39.36$\pm$1.14                             & \textbf{43.38$\pm$0.51}             \\ \hline
            stone\_brick     & {38.78$\pm$0.66} & \textbf{38.87$\pm$1.35}  & 38.76$\pm$1.11                      \\ \hline
            stone\_cement    & \textbf{42.32$\pm$1.26}    & 39.19$\pm$1.18       & 41.47$\pm$1.20                       \\ \hline
            turf             & 17.24$\pm$20.19   & \textbf{41.17$\pm$1.62}           & {37.61$\pm$16.88}            \\ \hline
            wood\_chips      & {\textbf{39.99$\pm$1.19}} & 38.76$\pm$1.12 & 39.64$\pm$1.66 \\ \hline
            All Classes      & 39.95$\pm$6.33  & 38.40$\pm$2.35  & \textbf{40.25$\pm$4.45}  \\ \hline
        \end{tabular}
		\caption{FDR Scores for remaining classes and overall}
		\label{tab:FDR_B}
		\end{subtable}
		\label{tab:FDR_scores}
	\end{table*}
	
	    \begin{figure*}[htb]
    \centering 
		\begin{subfigure}{.32\textwidth}{
				\includegraphics[draft=false,width=\textwidth]{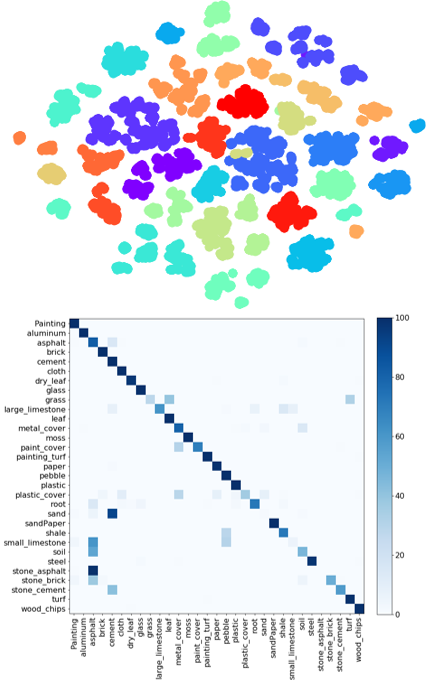}
				\caption{GAP (76.09$\pm$0.91\%)}
				\label{fig:GAP_fig}
			}
		\end{subfigure}
		\begin{subfigure}{.32\textwidth}{
				\includegraphics[draft=false,width=\textwidth]{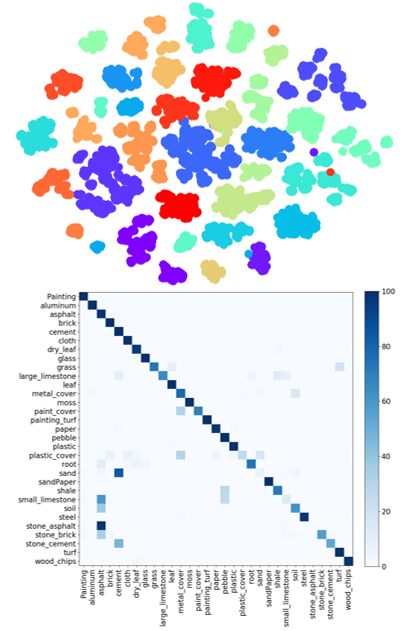}
				\caption{LinearHist\_$16$ (78.70$\pm$0.57\%)}
				\label{fig:Lin_fig}
			}
		\end{subfigure}
		\begin{subfigure}{.32\textwidth}{
				\includegraphics[draft=false,width=\textwidth]{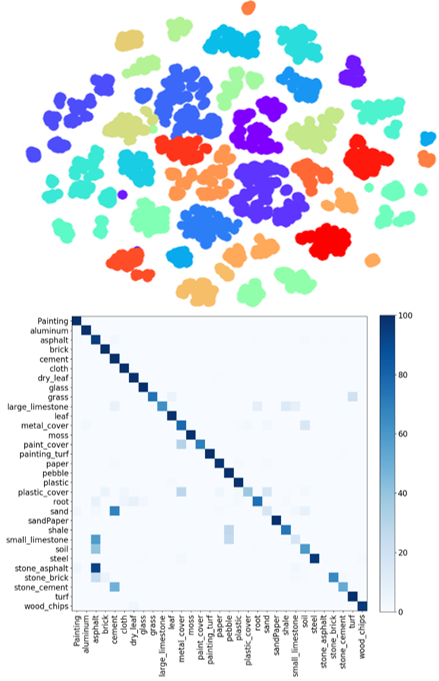}
				\caption{HistRes\_$16$ (ours, 79.75$\pm$0.84\%)}
				\label{fig:Hist_fig}
			}
		\end{subfigure}
		
		\caption{t-SNE \cite{maaten2008visualizing} 2-D visualization of features extracted before the fully connected layer from 10,000 randomly sampled training images from the GTOS-mobile dataset (the same random images were used for each model and random state for t-SNE) and the average test confusion matrices across the five runs of each model. The HistRes\_$16$ retains the compact clusters but also adds some separation between classes. The LinearHist\_$16$ achieves similar results but some classes appear to overlap more than our proposed model. The confusion matrices are colored based on class percentages. Overall, the histogram model improves the number of correctly classified textures, and it also reduces the number of misclassifications.}
		\centering
		\label{fig:CMTSNE_visual}
	\end{figure*}
	
	The histogram models for each dataset were compared to other state-of-the-art texture coding methods as well as the baseline ResNet models that only use GAP. The histogram models performed best for GTOS-mobile, but did not achieve the best performance for the other two datasets as shown in Table \mbox{\ref{tab:test_acc}}. As a result, the histogram layer may perform better in some applications than others (particularly for natural-looking texture datasets such as GTOS-mobile). Investigating a better initialization approach for the histogram that is similar to the synthetic dataset experiments (\textit{i.e.}, initialize equally spaced bins with equal width to cover feature value range) may also be beneficial to improve performance. We do not know the initial feature range of the images in the deep network. As a result, we randomly initialized the histogram parameters as described in Section \mbox{\ref{sect:architecture}} but future work would analyze alternative initialization approaches to optimize texture features extracted by the histogram layer.

	The binning functions used for each histogram model did not produce significant differences between one another, as far as for the overall accuracy. Since we are learning the bin centers and widths for the linear basis function and RBF, this shows that our model is fairly robust with regards to the selection of our approximation for the standard histogram operation. Similarly, each DTD model performed comparably, specifically when observing the performance for GAP* and the HistRes\_$B$ models. One reason for this is that a majority of the DTD dataset contains images with homogeneous textures and local information retained by the histogram layer; thus the model may not provide significant additional information. However, we found that, when compared to DeepTEN, a method that focuses on orderless texture information, the addition of spatial information provided by the proposed histogram layer provides improved performance. 
	
	For MINC-2500, most images have the texture of interest in local regions of the image. When looking at the results of the HistRest\_$B$ and GAP*, we observed that the addition of the histogram layer led to slightly worse performance. A possible reason for this is that the MINC-2500 dataset had several images that contained both background and the texture class. The context (\textit{i.e.}, region of image with texture of interest) varied throughout the dataset. As noted in \mbox{\cite{hu2019multi,xue2018deep,zhang2017deep}}, training the proposed models with 1) multiple images of different sizes and 2) varying scales in network would ideally lead to even more improved performance for our histogram layer in future work. The histogram model performed comparably to the other encoding approaches but again outperformed DeepTEN, further demonstrating the effectiveness of retaining spatial and texture information in the histogram layer. The DEP network also retained both spatial and texture information, but each HistRes\_$B$ achieved slightly better accuracy than the DEP model. MuLTER \cite{hu2019multi} incorporates texture at multiple levels, but our method also achieveed slightly better average accuracy as well and there was no significant difference between the results obtained.
	
	For GTOS-mobile, as shown in Table \mbox{\ref{tab:test_acc}}, our proposed method with the linear basis function or RBF binning operations achieves better performance in comparison to all other texture encoding methods as shown in Table \ref{tab:test_acc}. GTOS-mobile contains the most natural, textured images such as grass and sand. Histogram-based features, as shown throughout the literature \cite{liu2019bow,tuceryan1993texture}, effectively represent texture features within natural imagery. The textures are also located in various spatial regions throughout the imagery. As a result, our localized histogram models are able to capture this information to improve performance.
	
	In Figure \ref{fig:CMTSNE_visual}, the t-SNE visualization \cite{maaten2008visualizing} and confusion matrices are shown for the GAP and best histogram models for GTOS-mobile LinearHist\_16 and HistRes\_16). For the GAP model in Figure \ref{fig:GAP_fig}, the t-SNE visual shows that, while most classes in the training data are clustered well but some samples are not near their same class and there is some overlap between neighboring classes. In Figure \ref{fig:Hist_fig}, there is more separation between the classes. Also, samples that belong to the same class appear to be closer to one another. For the t-SNE visualizations of the linear basis function in Figure 
	6b, the classes appear to be more compact than the baseline model was. However, there are some samples that overlap in the projected space. Our RBF model retains compact and separated features for each class. A reason for this is that the linear basis function is a ``sharper" approximation of the standard histogram operation. The linear basis function will work well for some textures, but it will not be able to properly account for small intra-class variations of images. The RBF is a smoother approximation and will be better able to model small changes between samples of the same class. A future work would be to investigate heavier-tailed distributions (\textit{e.g.}, t-distribution) to analyze whether or not the model can account for larger intra-class variations. 
	
	To further validate our qualitative observations, we computed the log of Fisher's Discriminant Ratio (FDR) \cite{bishop2006pattern} to measure the ratio of separation and compactness for each class. For this, we used features from the 1) GAP, 2) LinearHist\_$16$, and 3) HistRes\_$16$. The quantitative values from the log of FDR matched our qualitative observation from the t-SNE visualization. The histogram features can be thought of as a similarity measure because a value between 0 (dissimilar) and 1 (similar) will be assigned to each bin based on centers and widths estimated by the model. If samples have similar textures, the distributions captured by the histogram layer should also be similar. Also, in the confusion matrices, both histogram models identified the ``sand" images, while the GAP model did not. The RBF model also classified some classes such as ``grass" and ``sand" slightly better than the linear basis function model. This also supports our previous hypothesis that the RBF is more robust to small intra-class varations since these two classes can appear visually different depending on the spatial arrangement of the texture. Overall, HistRes\_$16$ improved performance for most classes and reduced the number of misclassified textures. 
	
	\section{Conclusion}
	In this work, we presented a new layer for deep neural networks: a histogram layer. Previous deep learning approaches using CNNs are unable to effectively represent texture without adding more layers. Through the histogram layer, we directly captured statistical texture information within images by characterizing the distribution of feature maps extracted from CNN models. This meant that we could jointly fine-tune both the histogram layer and CNNs together.  Experiments on synthetic textures provided more insight into the functionality and utility of the proposed approach. We conducted an in-depth analysis of the configuration of the histogram layer, which provided more insight into the effects of normalization, sum-to-one constraints on the bins, and location of the histogram layer. When the HistRes\_$B$ models were compared to other state-of-the-art texture encoding approaches, they achieved improvements in performance, particularly for the most natural texture dataset, GTOS-mobile. The histogram layer can easily be integrated into other ANNs and used in such other texture analysis tasks besides classification such as segmentation, texture synthesis, and shape from texture \cite{liu2019bow}.


\ifCLASSOPTIONcaptionsoff
  \newpage
\fi


\printbibliography

@inproceedings{vaswani2017attention,
  title={Attention is all you need},
  author={Vaswani, Ashish and Shazeer, Noam and Parmar, Niki and Uszkoreit, Jakob and Jones, Llion and Gomez, Aidan N and Kaiser, {\L}ukasz and Polosukhin, Illia},
  booktitle={Advances in neural information processing systems},
  pages={5998--6008},
  year={2017}
}

@article{xu2018powerful,
  title={How powerful are graph neural networks?},
  author={Xu, Keyulu and Hu, Weihua and Leskovec, Jure and Jegelka, Stefanie},
  journal={arXiv preprint arXiv:1810.00826},
  year={2018}
}

@article{zaheer2017deep,
  title={Deep sets},
  author={Zaheer, Manzil and Kottur, Satwik and Ravanbakhsh, Siamak and Poczos, Barnabas and Salakhutdinov, Ruslan and Smola, Alexander},
  journal={arXiv preprint arXiv:1703.06114},
  year={2017}
}

@article{pinkus1999approximation,
  title={Approximation theory of the MLP model in neural networks},
  author={Pinkus, Allan},
  journal={Acta numerica},
  volume={8},
  pages={143--195},
  year={1999},
  publisher={Cambridge University Press}
}

@inproceedings{phuong2020inductive,
  title={The inductive bias of ReLU networks on orthogonally separable data},
  author={Phuong, Mary and Lampert, Christoph H},
  booktitle={International Conference on Learning Representations},
  year={2020}
}

@inproceedings{neyshabur2015search,
  title={In Search of the Real Inductive Bias: On the Role of Implicit Regularization in Deep Learning.},
  author={Neyshabur, Behnam and Tomioka, Ryota and Srebro, Nathan},
  booktitle={ICLR (Workshop)},
  year={2015}
}

@inproceedings{xu2021positional,
  title={Positional encoding as spatial inductive bias in gans},
  author={Xu, Rui and Wang, Xintao and Chen, Kai and Zhou, Bolei and Loy, Chen Change},
  booktitle={Proceedings of the IEEE/CVF Conference on Computer Vision and Pattern Recognition},
  pages={13569--13578},
  year={2021}
}

@article{wu2020comprehensive,
  title={A comprehensive survey on graph neural networks},
  author={Wu, Zonghan and Pan, Shirui and Chen, Fengwen and Long, Guodong and Zhang, Chengqi and Philip, S Yu},
  journal={IEEE transactions on neural networks and learning systems},
  volume={32},
  number={1},
  pages={4--24},
  year={2020},
  publisher={IEEE}
}

@article{bronstein2021geometric,
  title={Geometric deep learning: Grids, groups, graphs, geodesics, and gauges},
  author={Bronstein, Michael M and Bruna, Joan and Cohen, Taco and Veli{\v{c}}kovi{\'c}, Petar},
  journal={arXiv preprint arXiv:2104.13478},
  year={2021}
}

@inproceedings{zhu2021learning,
  title={Learning Statistical Texture for Semantic Segmentation},
  author={Zhu, Lanyun and Ji, Deyi and Zhu, Shiping and Gan, Weihao and Wu, Wei and Yan, Junjie},
  booktitle={Proceedings of the IEEE/CVF Conference on Computer Vision and Pattern Recognition},
  pages={12537--12546},
  year={2021}
}

@article{humeau2019texture,
	title={Texture feature extraction methods: A survey},
	author={Humeau-Heurtier, Anne},
	journal={IEEE Access},
	volume={7},
	pages={8975--9000},
	year={2019},
	publisher={IEEE}
}

@article{kang2021rotation,
  title={Rotation-Invariant deep embedding for remote sensing images},
  author={Kang, Jian and Fernandez-Beltran, Ruben and Wang, Zhirui and Sun, Xian and Ni, Jingen and Plaza, Antonio},
  journal={IEEE Transactions on Geoscience and Remote Sensing},
  year={2021},
  publisher={IEEE}
}

@inproceedings{laptev2016ti,
  title={Ti-pooling: transformation-invariant pooling for feature learning in convolutional neural networks},
  author={Laptev, Dmitry and Savinov, Nikolay and Buhmann, Joachim M and Pollefeys, Marc},
  booktitle={Proceedings of the IEEE conference on computer vision and pattern recognition},
  pages={289--297},
  year={2016}
}

@article{andrearczyk2018rotational,
  title={Rotational 3D texture classification using group equivariant CNNs},
  author={Andrearczyk, Vincent and Depeursinge, Adrien},
  journal={arXiv preprint arXiv:1810.06889},
  year={2018}
}

@inproceedings{marcos2016learning,
  title={Learning rotation invariant convolutional filters for texture classification},
  author={Marcos, Diego and Volpi, Michele and Tuia, Devis},
  booktitle={2016 23rd International Conference on Pattern Recognition (ICPR)},
  pages={2012--2017},
  year={2016},
  organization={IEEE}
}

@article{akhtar2019interpretation,
  title={Interpretation of intelligence in CNN-pooling processes: a methodological survey},
  author={Akhtar, Nadeem and Ragavendran, U},
  journal={Neural Computing and Applications},
  pages={1--20},
  publisher={Springer}
}

@inproceedings{hu2019multi,
  title={Multi-level texture encoding and representation (multer) based on deep neural networks},
  author={Hu, Yuting and Long, Zhiling and AlRegib, Ghassan},
  booktitle={2019 IEEE International Conference on Image Processing (ICIP)},
  pages={4410--4414},
  year={2019},
  organization={IEEE}
}

@book{bishop2006pattern,
  title={Pattern recognition and machine learning},
  author={Bishop, Christopher M},
  year={2006},
  publisher={springer}
}

@article{maaten2008visualizing,
	title={Visualizing data using t-SNE},
	author={Maaten, Laurens van der and Hinton, Geoffrey},
	journal={Journal of machine learning research},
	volume={9},
	number={Nov},
	pages={2579--2605},
	year={2008}
}

@inproceedings{abu2018robust,
	title={Robust Image Denoising for Sonar Imagery},
	author={Abu, Avi and Diamant, Roee},
	booktitle={2018 OCEANS-MTS/IEEE Kobe Techno-Oceans (OTO)},
	pages={1--5},
	year={2018},
	organization={IEEE}
}

@article{sedighi2017histogram,
	title={Histogram layer, moving convolutional neural networks towards feature-based steganalysis},
	author={Sedighi, Vahid and Fridrich, Jessica},
	journal={Electronic Imaging},
	volume={2017},
	number={7},
	pages={50--55},
	year={2017},
	publisher={Society for Imaging Science and Technology}
}

@inproceedings{wang2016learnable,
	title={Learnable histogram: Statistical context features for deep neural networks},
	author={Wang, Zhe and Li, Hongsheng and Ouyang, Wanli and Wang, Xiaogang},
	booktitle={European Conference on Computer Vision},
	pages={246--262},
	year={2016},
	organization={Springer}
}

@inproceedings{yang2010bag,
	title={Bag-of-visual-words and spatial extensions for land-use classification},
	author={Yang, Yi and Newsam, Shawn},
	booktitle={Proceedings of the 18th SIGSPATIAL international conference on advances in geographic information systems},
	pages={270--279},
	year={2010},
	organization={ACM}
}

@article{frigui2008detection,
	title={Detection and discrimination of land mines in ground-penetrating radar based on edge histogram descriptors and a possibilistic $ k $-nearest neighbor classifier},
	author={Frigui, Hichem and Gader, Paul},
	journal={IEEE Transactions on Fuzzy Systems},
	volume={17},
	number={1},
	pages={185--199},
	year={2008},
	publisher={IEEE}
}

@inproceedings{lowe1999object,
	title={Object recognition from local scale-invariant features.},
	author={Lowe, David G and others},
	booktitle={iccv},
	volume={99},
	number={2},
	pages={1150--1157},
	year={1999}
}

@inproceedings{dalal2005histograms,
	title={Histograms of oriented gradients for human detection},
	author={Dalal, Navneet and Triggs, Bill},
	booktitle={international Conference on computer vision \& Pattern Recognition (CVPR'05)},
	volume={1},
	pages={886--893},
	year={2005},
	organization={IEEE Computer Society}
}

@inproceedings{ojala1994performance,
	title={Performance evaluation of texture measures with classification based on Kullback discrimination of distributions},
	author={Ojala, Timo and Pietikainen, Matti and Harwood, David},
	booktitle={Proceedings of 12th International Conference on Pattern Recognition},
	volume={1},
	pages={582--585},
	year={1994},
	organization={IEEE}
}

@inproceedings{krizhevsky2012imagenet,
	title={Imagenet classification with deep convolutional neural networks},
	author={Krizhevsky, Alex and Sutskever, Ilya and Hinton, Geoffrey E},
	booktitle={Advances in neural information processing systems},
	pages={1097--1105},
	year={2012}
}

@inproceedings{he2016deep,
	title={Deep residual learning for image recognition},
	author={He, Kaiming and Zhang, Xiangyu and Ren, Shaoqing and Sun, Jian},
	booktitle={Proceedings of the IEEE conference on computer vision and pattern recognition},
	pages={770--778},
	year={2016}
}

@inproceedings{liang2015recurrent,
	title={Recurrent convolutional neural network for object recognition},
	author={Liang, Ming and Hu, Xiaolin},
	booktitle={Proceedings of the IEEE conference on computer vision and pattern recognition},
	pages={3367--3375},
	year={2015}
}

@inproceedings{long2015fully,
	title={Fully convolutional networks for semantic segmentation},
	author={Long, Jonathan and Shelhamer, Evan and Darrell, Trevor},
	booktitle={Proceedings of the IEEE conference on computer vision and pattern recognition},
	pages={3431--3440},
	year={2015}
}

@inproceedings{zhang2017deep,
	title={Deep ten: Texture encoding network},
	author={Zhang, Hang and Xue, Jia and Dana, Kristin},
	booktitle={Proceedings of the IEEE Conference on Computer Vision and Pattern Recognition},
	pages={708--717},
	year={2017}
}

@inproceedings{curio1999walking,
	title={Walking pedestrian recognition},
	author={Curio, Cristobal and Edelbrunner, Johann and Kalinke, Thomas and Tzomakas, Christos and Von Seelen, Werner},
	booktitle={Proceedings 199 IEEE/IEEJ/JSAI International Conference on Intelligent Transportation Systems (Cat. No. 99TH8383)},
	pages={292--297},
	year={1999},
	organization={IEEE}
}

@article{castellano2004texture,
	title={Texture analysis of medical images},
	author={Castellano, G and Bonilha, L and Li, LM and Cendes, F},
	journal={Clinical radiology},
	volume={59},
	number={12},
	pages={1061--1069},
	year={2004},
	publisher={Elsevier}
}

@article{anderson2012combination,
	title={Combination of anomaly algorithms and image features for explosive hazard detection in forward looking infrared imagery},
	author={Anderson, Derek T and Stone, Kevin E and Keller, James M and Spain, Christopher J},
	journal={IEEE Journal of Selected Topics in Applied Earth Observations and Remote Sensing},
	volume={5},
	number={1},
	pages={313--323},
	year={2012},
	publisher={IEEE}
}

@article{basu2018deep,
	title={Deep neural networks for texture classification-A theoretical analysis},
	author={Basu, Saikat and Mukhopadhyay, Supratik and Karki, Manohar and DiBiano, Robert and Ganguly, Sangram and Nemani, Ramakrishna and Gayaka, Shreekant},
	journal={Neural Networks},
	volume={97},
	pages={173--182},
	year={2018},
	publisher={Elsevier}
}

@inproceedings{basu2016theoretical,
	title={A theoretical analysis of deep neural networks for texture classification},
	author={Basu, Saikat and Karki, Manohar and Mukhopadhyay, Supratik and Ganguly, Sangram and Nemani, Ramakrishna and DiBiano, Robert and Gayaka, Shreekant},
	booktitle={2016 International Joint Conference on Neural Networks (IJCNN)},
	pages={992--999},
	year={2016},
	organization={IEEE}
}

@inproceedings{cavalin2017review,
	title={A review of texture classification methods and databases},
	author={Cavalin, Paulo and Oliveira, Luiz S},
	booktitle={2017 30th SIBGRAPI Conference on Graphics, Patterns and Images Tutorials (SIBGRAPI-T)},
	pages={1--8},
	year={2017},
	organization={IEEE}
}

@article{liu2017local,
	title={Local binary features for texture classification: Taxonomy and experimental study},
	author={Liu, Li and Fieguth, Paul and Guo, Yulan and Wang, Xiaogang and Pietik{\"a}inen, Matti},
	journal={Pattern Recognition},
	volume={62},
	pages={135--160},
	year={2017},
	publisher={Elsevier}
}

@book{Goodfellow-et-al-2016,
	title={Deep Learning},
	author={Ian Goodfellow and Yoshua Bengio and Aaron Courville},
	publisher={MIT Press},
	note={\url{http://www.deeplearningbook.org}},
	year={2016}
}

@article{holub2013random,
	title={Random projections of residuals for digital image steganalysis},
	author={Holub, Vojtech and Fridrich, Jessica},
	journal={IEEE Transactions on Information Forensics and Security},
	volume={8},
	number={12},
	pages={1996--2006},
	year={2013},
	publisher={IEEE}
}

@article{ojala2004texture,
	title={Texture Classification, Machine Vision, and Media Processing Unit},
	author={Ojala, T and Pietik{\"a}inen, M},
	journal={University of Oulu, Finland},
	year={2004}
}

@inproceedings{srinivasan2008statistical,
	title={Statistical texture analysis},
	author={Srinivasan, GN and Shobha, G},
	booktitle={Proceedings of world academy of science, engineering and technology},
	volume={36},
	pages={1264--1269},
	year={2008}
}

@article{haralick1973textural,
	title={Textural features for image classification},
	author={Haralick, Robert M and Shanmugam, Karthikeyan and others},
	journal={IEEE Transactions on systems, man, and cybernetics},
	number={6},
	pages={610--621},
	year={1973},
	publisher={Ieee}
}

@article{ojala2002multiresolution,
	title={Multiresolution gray-scale and rotation invariant texture classification with local binary patterns},
	author={Ojala, Timo and Pietik{\"a}inen, Matti and M{\"a}enp{\"a}{\"a}, Topi},
	journal={IEEE Transactions on Pattern Analysis \& Machine Intelligence},
	number={7},
	pages={971--987},
	year={2002},
	publisher={IEEE}
}

@article{liu2019bow,
	title={From BoW to CNN: Two decades of texture representation for texture classification},
	author={Liu, Li and Chen, Jie and Fieguth, Paul and Zhao, Guoying and Chellappa, Rama and Pietik{\"a}inen, Matti},
	journal={International Journal of Computer Vision},
	volume={127},
	number={1},
	pages={74--109},
	year={2019},
	publisher={Springer}
}

@inproceedings{paul2016combining,
	title={Combining deep neural network and traditional image features to improve survival prediction accuracy for lung cancer patients from diagnostic CT},
	author={Paul, Rahul and Hawkins, Samuel H and Hall, Lawrence O and Goldgof, Dmitry B and Gillies, Robert J},
	booktitle={2016 IEEE International Conference on Systems, Man, and Cybernetics (SMC)},
	pages={002570--002575},
	year={2016},
	organization={IEEE}
}

@inproceedings{wu2016multi,
	title={Multi-view common space learning for emotion recognition in the wild},
	author={Wu, Jianlong and Lin, Zhouchen and Zha, Hongbin},
	booktitle={Proceedings of the 18th ACM International Conference on Multimodal Interaction},
	pages={464--471},
	year={2016},
	organization={ACM}
}

@article{liu2019texture,
	title={Texture Classification in Extreme Scale Variations using GANet},
	author={Liu, Li and Chen, Jie and Zhao, Guoying and Fieguth, Paul and Chen, Xilin and Pietik{\"a}inen, Matti},
	journal={IEEE Transactions on Image Processing},
	year={2019},
	publisher={IEEE}
}

@article{wang2014mitosis,
	title={Mitosis detection in breast cancer pathology images by combining handcrafted and convolutional neural network features},
	author={Wang, Haibo and Roa, Angel Cruz and Basavanhally, Ajay N and Gilmore, Hannah L and Shih, Natalie and Feldman, Mike and Tomaszewski, John and Gonzalez, Fabio and Madabhushi, Anant},
	journal={Journal of Medical Imaging},
	volume={1},
	number={3},
	pages={034003},
	year={2014},
	publisher={International Society for Optics and Photonics}
}

@article{bruna2013invariant,
	title={Invariant scattering convolution networks},
	author={Bruna, Joan and Mallat, St{\'e}phane},
	journal={IEEE transactions on pattern analysis and machine intelligence},
	volume={35},
	number={8},
	pages={1872--1886},
	year={2013},
	publisher={IEEE}
}

@article{chan2015pcanet,
	title={PCANet: A simple deep learning baseline for image classification?},
	author={Chan, Tsung-Han and Jia, Kui and Gao, Shenghua and Lu, Jiwen and Zeng, Zinan and Ma, Yi},
	journal={IEEE transactions on image processing},
	volume={24},
	number={12},
	pages={5017--5032},
	year={2015},
	publisher={IEEE}
}

@inproceedings{malof2018improving,
	title={Improving the histogram of oriented gradient feature for threat detection in ground penetrating radar by implementing it as a trainable convolutional neural network},
	author={Malof, Jordan M and Bralich, John and Reichman, Dani{\"e}l and Collins, Leslie M},
	booktitle={Detection and Sensing of Mines, Explosive Objects, and Obscured Targets XXIII},
	volume={10628},
	pages={106280D},
	year={2018},
	organization={International Society for Optics and Photonics}
}

@inproceedings{song2017locally,
	title={Locally-transferred fisher vectors for texture classification},
	author={Song, Yang and Zhang, Fan and Li, Qing and Huang, Heng and O'Donnell, Lauren J and Cai, Weidong},
	booktitle={Proceedings of the IEEE International Conference on Computer Vision},
	pages={4912--4920},
	year={2017}
}

@inproceedings{rivera2018densenet,
	title={DenseNet model combined with Haralick’s handcrafted features for texture classification},
	author={Rivera-Morales, Carlos-Andres and Bastidas-Rodr{\'\i}guez, Maria-Ximena and Prieto-Ortiz, Flavio-Augusto},
	booktitle={2018 IEEE Latin American Conference on Computational Intelligence (LA-CCI)},
	pages={1--6},
	year={2018},
	organization={IEEE}
}

@inproceedings{xue2018deep,
	title={Deep texture manifold for ground terrain recognition},
	author={Xue, Jia and Zhang, Hang and Dana, Kristin},
	booktitle={Proceedings of the IEEE Conference on Computer Vision and Pattern Recognition},
	pages={558--567},
	year={2018}
}

@inproceedings{liu2016evaluation,
	title={Evaluation of LBP and deep texture descriptors with a new robustness benchmark},
	author={Liu, Li and Fieguth, Paul and Wang, Xiaogang and Pietik{\"a}inen, Matti and Hu, Dewen},
	booktitle={European Conference on Computer Vision},
	pages={69--86},
	year={2016},
	organization={Springer}
}

@inproceedings{cimpoi2015deep,
	title={Deep filter banks for texture recognition and segmentation},
	author={Cimpoi, Mircea and Maji, Subhransu and Vedaldi, Andrea},
	booktitle={Proceedings of the IEEE conference on computer vision and pattern recognition},
	pages={3828--3836},
	year={2015}
}

@InProceedings{cimpoi14describing,
	Author    = {M. Cimpoi and S. Maji and I. Kokkinos and S. Mohamed and and A. Vedaldi},
	Title     = {Describing Textures in the Wild},
	Booktitle = {Proceedings of the {IEEE} Conf. on Computer Vision and Pattern Recognition ({CVPR})},
	Year      = {2014}}

@inproceedings{bell2015material,
	title={Material recognition in the wild with the materials in context database},
	author={Bell, Sean and Upchurch, Paul and Snavely, Noah and Bala, Kavita},
	booktitle={Proceedings of the IEEE conference on computer vision and pattern recognition},
	pages={3479--3487},
	year={2015}
}

@inproceedings{yu2014mixed,
	title={Mixed pooling for convolutional neural networks},
	author={Yu, Dingjun and Wang, Hanli and Chen, Peiqiu and Wei, Zhihua},
	booktitle={International Conference on Rough Sets and Knowledge Technology},
	pages={364--375},
	year={2014},
	organization={Springer}
}

@incollection{tuceryan1993texture,
	title={Texture analysis},
	author={Tuceryan, Mihran and Jain, Anil K},
	booktitle={Handbook of pattern recognition and computer vision},
	pages={235--276},
	year={1993},
	publisher={World Scientific}
}




\end{document}